\documentclass[11pt]{article}

\usepackage[preprint]{acl}

\usepackage{times}
\usepackage{latexsym}

\usepackage[T1]{fontenc}
\usepackage[utf8]{inputenc}

\usepackage{microtype}

\usepackage{inconsolata}

\usepackage{graphicx}

\usepackage{url}
\usepackage{xurl}
\usepackage{booktabs}
\usepackage{amsmath}
\usepackage{amsfonts}
\usepackage{amssymb}
\usepackage{amsthm}
\usepackage{nicefrac}
\usepackage{xcolor}
\usepackage{multirow}
\usepackage{makecell}
\usepackage{float}
\usepackage{enumitem}

\newtheorem{proposition}{Proposition}
\newtheorem{example}{Example}

\newcommand{\revcolor}{black}
\newcommand{\rev}[1]{{\color{\revcolor}#1}}

\title{A Simple Plug-in for Improving \\ Eviction-Based KV Cache Compression}

\author{
  \textbf{Yuping Lin\textsuperscript{1}}\thanks{Work done during internship at Hippocratic AI.},
  \textbf{Jiayuan Ding\textsuperscript{2}},
  \textbf{Yue Xing\textsuperscript{1}},
  \textbf{Pengfei He\textsuperscript{1}},
\\
  \textbf{Jiliang Tang\textsuperscript{1}},
  \textbf{Subhabrata Mukherjee\textsuperscript{2}}
\\
\\
  \textsuperscript{1}Michigan State University,
  \textsuperscript{2}Hippocratic AI
\\
  \small{\texttt{\{linyupin,xingyue1,hepengf1,tangjili\}@msu.edu}}
\\
  \small{\texttt{\{jiayuan,subho\}@hippocraticai.com}}
}

\begin{document}
\maketitle
\begin{abstract}
KV cache growth is a major bottleneck for long-context inference in large language models. Existing methods are often dominated by binary eviction or representation approximation, which may underutilize tokens that are not critical for exact retention but are still reconstructable. We present VECTOR, a plug-and-play augmentation for eviction-based pipelines that introduces three-way token routing: retention, approximation, and eviction. VECTOR combines an importance signal from the base scorer with a reconstructability signal from an offline-calibrated regression-based value estimation. By leveraging reconstructability, VECTOR recovers useful value information that would otherwise be irreversibly lost under binary eviction, while preserving key vectors for attention routing stability. 
% [REV-DEL] Experimental results show that VECTOR improves quality-memory trade-offs under medium-to-high compression, with especially clear gains in stricter budget regimes.
\rev{On LongBench, RULER with contexts up to 128K tokens, and multi-turn SCBench, VECTOR yields significant gains over its base eviction methods under high compression. At a 90\% compression ratio, SnapKV augmented with VECTOR outperforms six competing compression methods, including D2O and RocketKV, on all three evaluated models. Since VECTOR makes all routing decisions at prefill, decoding runs on a standard compressed cache with no measurable overhead.}
\end{abstract}

\section{Introduction}\label{sec:introduction}

Large language models (LLMs) increasingly rely on long-context inference~\citep{grattafiori2024llama, team2024gemini, singh2025openai}, yet their key-value (KV) cache grows linearly with sequence length and quickly becomes the dominant memory cost~\citep{kwon2023efficient,zhang2023h2o}. This bottleneck limits practical deployment in many applications such as retrieval-heavy QA~\citep{lewis2020retrieval}, agent workflows~\citep{yao2022react}, and multi-turn reasoning~\rev{\citep{packer2023memgpt, li2025scbench}}, where context windows are long and memory budgets are strict. As a result, KV cache compression has become a central system problem for long-context LLM serving~\citep{kwon2023efficient}.

Existing KV compression methods mainly follow two strategies. The first strategy is \emph{importance-based eviction}, where tokens are scored and low-score entries are permanently evicted (e.g., SnapKV~\citep{li2024snapkv}, KeyDiff~\citep{park2025keydiff}, KVzip~\citep{kim2025kvzip}). While these methods are simple and efficient, their decision is inherently binary: tokens are either fully retained or completely discarded. Under tight budgets, such an irreversible eviction can substantially degrade downstream performance\rev{. Recent methods move beyond this binary decision along two axes. D2O~\citep{wan2024d2o} works on the token axis with retain, merge, and evict decisions, while ARKV~\citep{lei2026arkv} works on the precision axis with full-precision, quantized, and evicted states.}
% [REV-DEL] , especially when recent multi-state systems such as ARKV~\citep{lei2026arkv} and D2O~\citep{wan2024d2o} suggest that non-binary allocation can outperform pure eviction.
% because binary eviction permanently removes tokens that could otherwise be approximately reconstructed.

% \yp{TODO: rewrite from here..}

The second strategy is \emph{representation approximation}, which compresses KV representations through quantization, projection, or reconstruction (e.g., AQUA-KV~\citep{shutova2025cache}, EliteKV~\citep{zhou2025elitekv}, DeltaKV~\citep{hao2026deltakv}, Attention Matching~\citep{zweiger2026fast}). Rather than removing tokens outright, these methods attempt to preserve information approximately after compression. However, many existing approaches require architectural modifications, retraining, or expensive online computation, such as retrieval-based reconstruction~\citep{hao2026deltakv}, context-dependent fitting~\citep{zweiger2026fast}, or model uptraining~\citep{zhou2025elitekv}.

To better balance memory reduction and downstream performance, we treat KV compression as a unified \emph{retain–approximate–evict} allocation problem, where tokens are selectively retained, approximately reconstructed, or removed under a fixed memory budget. This perspective is motivated by two observations. First, recent \rev{non-binary} methods~\citep{lei2026arkv, wan2024d2o} demonstrate that \rev{going beyond binary eviction improves performance}.
% [REV-DEL] First, recent multi-state methods~\citep{lei2026arkv, wan2024d2o} demonstrate that non-binary allocation outperforms pure eviction.
 However, their allocation decisions are driven by token importance signals such as attention scores or attention statistics. They do not explicitly model \emph{reconstructability}, namely, whether a token's KV representation can be accurately recovered from other available information with limited error. This distinction matters because not all important tokens are equally compressible: some can be safely approximated, while others are highly error-sensitive and should instead be retained exactly. Second, prior work on KV quantization suggests that keys and values exhibit asymmetric sensitivity to compression error, with values generally being more tolerant to approximation than keys~\citep{liu2024kivi, li2025kvtuner}. However, existing token-level KV compression methods rarely exploit this asymmetry explicitly when allocating memory across tokens. \rev{This leaves room to improve both performance and compression efficiency.}

% While the above two strategies are useful, two pieces remain underexplored. First, although existing methods consider \textit{token importance} or use \textit{representation approximation}, they do not explicitly differentiate tokens based on \textit{reconstructability}, i.e., whether a token's representation can be approximated from other available information with limited error (measured by reconstruction error in our method). Second, prior work on asymmetric KV quantization suggests that Keys and Values have different error sensitivity in practice~\citep{liu2024kivi, li2025kvtuner}. However, existing KV compression methods rarely exploit this asymmetry explicitly in token-level allocation.

% \yp{... to here. Logic: previous methods either 1) modify model arch or retrain; 2. heavy online compute; 3. threeway but only importance; 4. underexplore KV asymmetry. (be careful about 4)}

% These two gaps indicate that KV compression should not be treated as a uniform operation: the decisions of \textit{which} tokens to retain and \textit{how} to represent their KV representations should be optimized jointly.

% \jt{it is not clear to me that how the following two research questions related to the above challenges? how our answers to these questions address the above challenges??? }

% Motivated by these challenges, we raise two research questions:
Motivated by these \rev{observations}, we raise two research questions. The first concerns the approximation mechanism; the second concerns the allocation strategy.
\begin{itemize}[leftmargin=*,itemsep=1pt,topsep=3pt]
    \item \textbf{RQ1}: Can we build a lightweight reconstruction mechanism that recovers value representations with low runtime overhead and no architectural retraining?
    \item \textbf{RQ2}: Under a fixed KV memory budget, how should we jointly use \emph{importance} and \emph{reconstructability} to allocate tokens into \emph{retention}, \emph{approximation}, and \emph{eviction} tiers?
\end{itemize}

To answer these questions, we propose \textbf{VECTOR} (\textbf{V}alue \textbf{E}stimation via \textbf{C}ollinearity and \textbf{T}hree-way \textbf{O}rthogonal \textbf{R}outing), a plug-and-play augmentation for token-importance-based eviction methods. 
% VECTOR is \emph{not} a standalone replacement for existing scorers. Instead, it extends the binary decision into a three-way allocation where tokens are \emph{retained}, \emph{approximated}, or \emph{evicted}.
% [REV-DEL] {Unlike multi-state methods such as ARKV~\citep{lei2026arkv} and D2O~\citep{wan2024d2o}, which are standalone compression pipelines driven by importance signals alone, VECTOR augments existing eviction backbones by introducing reconstructability as a second allocation dimension.}
\rev{Unlike D2O~\citep{wan2024d2o}, a standalone token-level pipeline driven by importance signals alone, VECTOR augments existing eviction backbones by introducing reconstructability as a second allocation dimension. VECTOR acts on the token axis, so it is orthogonal to precision-level methods such as ARKV~\citep{lei2026arkv}.}

\begin{figure}[t]
    \centering
    \includegraphics[width=\columnwidth]{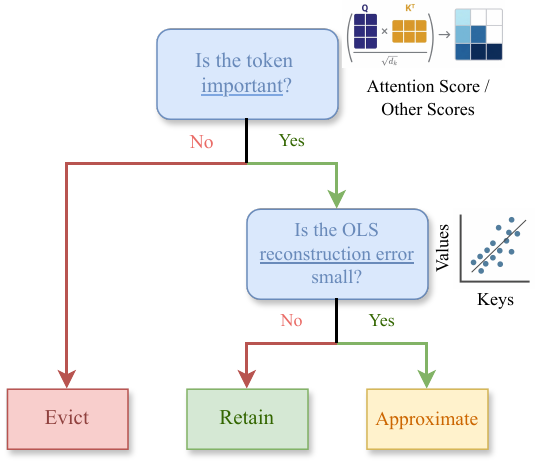}
    \caption{Overview of VECTOR's three-way token allocation. The base importance scorer first filters out unimportant tokens for eviction. For important tokens, VECTOR evaluates OLS-based $K\!\to\!V$ reconstruction error: tokens with small error enter \textit{Approximation}, while tokens with large error remain in \textit{Retention}. Approximation is applied to values only (V-only), with keys retained.}
    \label{fig:threeway-illustration}
\vspace{-1.0em}% [REV-VSPACE]
\end{figure}

\rev{For \textbf{RQ1}, VECTOR reconstructs values from keys with an offline-calibrated Ordinary Least Squares (OLS) map. Keys and values are linear projections of the same hidden state, which makes linear prediction feasible. VECTOR keeps keys exactly because they determine the softmax weights, where errors are amplified, while values are aggregated linearly and tolerate approximation~\citep{liu2024kivi,li2025kvtuner}.}

% [REV-DEL] For \textbf{RQ2}, VECTOR performs budgeted three-way allocation on two orthogonal axes: token \emph{importance} (from a base eviction policy) and token \emph{reconstructability} (from OLS reconstruction error). Intuitively, hard-to-reconstruct tokens should be retained, easy-to-reconstruct tokens can be approximated, and low-importance tokens can be evicted. Under the same KV memory budget, this reconstructability-aware allocation recovers part of the information that binary eviction would discard, enabling better quality-memory trade-offs under high compression.
\rev{For \textbf{RQ2}, VECTOR allocates tokens along two orthogonal axes: \emph{importance} from the base eviction method, and \emph{reconstructability} from the OLS reconstruction error. Low-importance tokens are evicted. Among the rest, easy-to-reconstruct tokens are approximated, and the others are retained exactly. Under the same memory budget, this recovers information that binary eviction would discard.}

Our main contributions are summarized as follows:

\begin{itemize}[leftmargin=*,itemsep=1pt,topsep=3pt]
    \item \textbf{A reconstructability-aware allocation view of KV compression.} As in Figure \ref{fig:threeway-illustration}, we formulate KV budget allocation as a three-way token allocation problem that jointly considers \textit{importance} and \textit{reconstructability}, rather than a binary \textit{retention}/\textit{eviction} decision.

    \item \textbf{A lightweight K$\rightarrow$V reconstruction module.} We develop a one-time offline OLS calibration that enables online value reconstruction from the stored keys, with low inference-time overhead.

    \item \textbf{A plug-and-play upgrade for eviction baselines.} VECTOR can be attached to representative token-importance eviction methods (e.g., attention-score-based and key-geometry-based) with minimal integration adaptations.

% [REV-DEL]     \item \textbf{Empirical gains under strict memory budgets.} Experiments on long-context benchmarks show that adding VECTOR to existing eviction baselines improves downstream performance in high-compression regimes across our evaluated settings.
    \item \rev{\textbf{Empirical gains under strict memory budgets.} On LongBench, RULER with up to 128K tokens, and multi-turn SCBench, adding VECTOR to eviction baselines yields significant gains under high compression. At $p_c=0.90$, SnapKV+VECTOR outperforms six competing compression methods on all three evaluated models, and VECTOR adds no measurable decoding overhead.}
\end{itemize}

\section{Related Work}

\paragraph{KV Cache Compression}
Existing methods fall into three families. Eviction-based methods score tokens by importance and permanently discard low-scoring entries~\rev{\citep{zhang2023h2o, li2024snapkv, park2025keydiff, cai2024pyramidkv, kim2025kvzip, oren2024tova, feng2024adakv, liu2025chunkkv}}. They are effective but operate as binary keep-or-drop policies, causing irreversible information loss under tight budgets. Quantization methods reduce bit-width across all cached entries~\citep{liu2024kivi, kang2024gear, li2025kvtuner, shutova2025cache}. They are less destructive but bounded in compression ratio and often require hardware support. Approximation methods reconstruct KV representations through low-rank projection or residual modeling~\citep{chang2025palu, hao2026deltakv, zweiger2026fast, zhou2025elitekv}. % [REV-DEL] Recent multi-state designs such as ARKV and D2O show that non-binary allocation increasingly matters under aggressive compression~\citep{lei2026arkv, wan2024d2o}.
\rev{Non-binary allocation is also increasingly common, along two distinct axes. On the token axis, D2O uses retain, merge, and evict decisions~\citep{wan2024d2o}. On the precision axis, ARKV assigns full-precision, quantized, and evicted states~\citep{lei2026arkv}. Another line keeps a larger KV pool and selects tokens dynamically during decoding, through sparse attention~\citep{behnam2025rocketkv} or CPU offloading~\citep{sun2025shadowkv}.}

\paragraph{Exploiting K--V Subspace Redundancy}
Since K and V are both projected from the same hidden state, their shared structure can be exploited for compression. MLA compresses token memory into a shared latent representation~\citep{liu2024deepseek}. AQUA-KV and EliteKV leverage K/V dependency through residual quantization and joint low-rank projection, respectively~\citep{shutova2025cache,zhou2025elitekv}. These works support the view that \rev{the shared structure of K and V} is a practically exploitable property.

\paragraph{Relation to Our Method}
VECTOR is a plug-and-play augmentation for eviction-based pipelines, not a standalone compression method. Unlike approximation-oriented approaches~\citep{chang2025palu,hao2026deltakv}, it requires no architectural modification or retraining and preserves full compatibility with existing eviction backbones. Within the eviction family, VECTOR extends binary retain/evict decisions by introducing a third approximation tier driven by reconstructability, a dimension that prior \rev{non-binary} methods~\citep{lei2026arkv, wan2024d2o} do not explicitly model. \rev{Because VECTOR acts on the token axis, it is orthogonal to precision-based compression.} We therefore evaluate VECTOR as an augmentation over eviction baselines, which directly reflects its intended use case\rev{, and further compare it with six competing compression methods (Section~\ref{sec:experiments:competing})}.
\section{Method}\label{sec:method}

% \jt{I do not get the logic of the following organization of subsections. why Approximation first, how Approximation related to Vector?? I think it will be better to leverage figure 1 to high-levely intro VECTOR  }

% \yue{Motivated by the limitations in existing KV cache compression algorithms in Section \ref{sec:introduction}, in this section, we provide details for the proposed algorithm.

% In our framework, besides the \textit{importance} score (e.g., attention scores \cite{li2024snapkv} or geometric features in Key space \cite{park2025keydiff}), we introduce a second dimension, \textit{reconstructability}. We define a token as highly reconstructable if its KV representation can be predicted accurately from other available information. By jointly modeling \textit{importance} and \textit{reconstructability}, we generalize binary retain/evict decisions to a three-way allocation: \textit{retention}, \textit{approximation}, and \textit{eviction}. 
% % In our framework, the \textit{approximation} tier is determined by both dimensions, enabling finer-grained and more effective budget allocation than importance-only policies.

% In the following,  the approximation approach is detailed in Section \ref{sec:approximation}, and the overall pipeline is in Section \ref{sec:method_pipeline}.
% }

VECTOR extends token-importance-based eviction \rev{methods} by introducing a third allocation state: \textit{approximation}. As in Figure~\ref{fig:threeway-illustration}, token routing proceeds along two dimensions. The first is \textit{importance}, provided by a base eviction scorer. The second is \textit{reconstructability}, which measures how accurately a token's value can be recovered from its key via a lightweight linear predictor. Together, these two dimensions determine whether each token is retained exactly, approximated, or evicted.

\subsection{\textit{Approximation}: Value Estimation via K--V Collinearity}\label{sec:approximation}

While the above discussion motivates the incorporation of \textit{reconstructability} into KV cache compression, it is crucial to recognize that keys and values exhibit different tolerances to approximation errors. Recent studies \citep{li2025kvtuner, patel2026turboangle} have shown that even small perturbations in key vectors are exponentially amplified by the softmax function, leading to catastrophic disruptions in the attention probability distribution. In contrast, errors in value vectors propagate only linearly through the weighted sum, as values are applied after the softmax normalization. This asymmetry suggests that approximating keys is significantly more risky than approximating values.

Building on this observation, we adopt a \textbf{$V$-only estimation} strategy for tokens in the \textit{Approximation} state. Specifically, we retain the exact key vector to preserve the integrity of the attention routing mechanism, while discarding the original value vector to reclaim memory. The missing value is subsequently reconstructed using an alternative compact representation. This asymmetric treatment ensures that the critical attention map remains uncompromised while maximizing memory savings. 
We empirically validate this design choice in Appendix~\ref{app:vonly_vs_konly}, where K-only approximation shows inconsistent gains and occasionally degrades below the unaugmented baseline under medium-to-high compression, while V-only approximation consistently improves over it.

% [REV-DEL] To validate this approach, we provide an empirical evidence in the following.
\rev{We next verify this approach empirically.}

\begin{table}[t]
    \centering
    \small
    \begin{tabular}{lc}
        \toprule
        Model & $R^2_\textrm{global}$ (All-Layer Avg.) \\
        \midrule
        Llama-3.1-8B & 0.6964 \\
        Qwen3-14B & 0.6946 \\
        Qwen3-0.6B & 0.9392 \\
        Gemma-3-4B & 0.8896 \\
        Qwen3-30B-A3B & 0.6863 \\
        \bottomrule
    \end{tabular}
    \caption{Cross-model $K \rightarrow V$ predictability with offline OLS. We report layer-averaged $R^2_\textrm{global}$ for each model, showing high linear predictability across model families, scales, and architectures.}
    \label{tab:method_r2}
\vspace{-1.0em}% [REV-VSPACE]
\end{table}
\paragraph{Empirical Evidence} Before presenting the full method design, we first verify a prerequisite of our approach: \textit{whether a simple linear map can reliably predict $V$ from $K$ across different model families, scales, and architectures.} We employ the C4 dataset~\cite{dodge2021documenting}, a large-scale pre-training corpus, to collect key and value activation vectors at each transformer layer, drawing from 10,000 sequences of 4,096 tokens. For each layer, a linear model predicting $V$ from $K$ is subsequently fitted via OLS regression. To evaluate generalization, we additionally construct a held-out test set consisting of 100 sequences of 4,096 tokens, and report the mean $R^2$ of the OLS averaged across all layers within each model on this test set. As shown in Table~\ref{tab:method_r2}, offline OLS achieves high $K \rightarrow V$ predictability on all tested models, with $R^2_\textrm{global}$ ranging from $0.6863$ to $0.9392$. This result serves as the empirical foundation of our approximation strategy.

\rev{\paragraph{Remark} In standard attention, both $K \in \mathbb{R}^{d_k}$ and $V \in \mathbb{R}^{d_v}$ are linear projections of the same hidden state $h \in \mathbb{R}^d$, i.e., $K = W_Kh$ and $V = W_Vh$.\footnote{Throughout this paper, we denote all vectors as column vectors, and linear transformations are written as $y = Wx$.} Exact reconstruction of $V$ from $K$ is impossible in general, since modern architectures use $d_k, d_v \ll d$~\citep{vaswani2017attention, shazeer2019fast, ainslie2023gqa, grattafiori2024llama}. However, the variance of $h$ is dominated by a low effective rank~\citep{sun2024massive}, and Multi-Head Latent Attention~\citep{liu2024deepseek} shows that $K$ and $V$ can share a low-dimensional latent representation. Both $K$ and $V$ are therefore well approximated by projections of a shared low-dimensional subspace, which makes predicting $V$ from $K$ feasible.}

\rev{\paragraph{Why Not the Pseudo-Inverse?} Since $K=W_Kh$ and $V=W_Vh$, a calibration-free alternative is $\hat{V}=W_V W_\mathrm{MP} K$, where $W_\mathrm{MP} = W_K^\top(W_K W_K^\top)^{-1}$ is the Moore-Penrose pseudo-inverse of $W_K$. This estimator picks the minimum-norm $h'$ consistent with $K$, which is not chosen to predict $V$ well. In practice, it gives negative $R^2$ on most layers of four of the five models in Table~\ref{tab:method_r2}, while OLS reaches 0.69 to 0.94 (Appendix~\ref{app:ols_vs_mp}). We therefore fit $W_\mathrm{OLS} = \arg\min_{W} \mathbb{E}_{h \sim \mathcal{D}} \|WK - V\|_2^2$ to minimize the $V$-prediction error directly, where $\mathcal{D}$ is the distribution of hidden states on a pre-training corpus. The calibration is also cheap. On Llama-3.1-8B, calibrating on $10\times$ fewer sequences (500) or on a different corpus does not significantly change downstream results (Appendix~\ref{app:calib}).}

\paragraph{RoPE Decoupling} A critical engineering challenge arises with Rotary Position Embedding (RoPE)~\citep{su2024roformer}, which is ubiquitously adopted in modern LLMs. RoPE applies a position-dependent rotation to the keys ($K_\textrm{post-RoPE} = R_m W_K h$ at position $m$), while values remain position-independent ($V = W_V h$). Fitting a static OLS matrix directly on post-RoPE keys would necessitate position-dependent estimators due to the entanglement of positional information, adding unnecessary complexity. To expose the intrinsic K-V collinearity, we explicitly decouple the positional information by applying an inexpensive inverse rotation ($R_m^{-1} K_\textrm{post-RoPE}$) to the cached keys during the approximation phase. This lightweight decoupling allows the static $W_\textrm{OLS}$ to accurately reconstruct $V$ regardless of token position.

\subsection{The VECTOR Pipeline}
\label{sec:method_pipeline}

\begin{figure}[t]
    \centering
    \includegraphics[width=0.85\columnwidth]{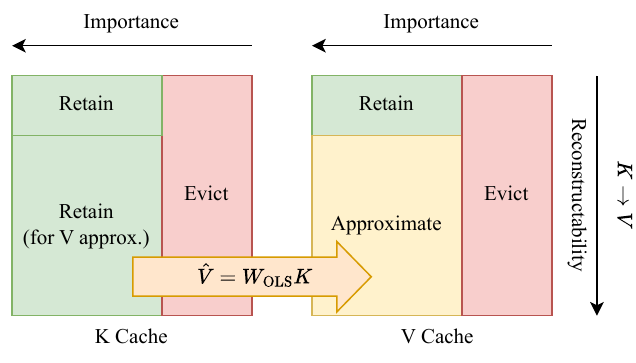}
    \caption{Asymmetric three-way allocation: K is retained exactly for the expanded candidate pool, while V is split into \textit{Retain}, \textit{Approximate}, and \textit{Evict}.}
    \label{fig:orthogonal}
\vspace{-1.0em}% [REV-VSPACE]
\end{figure}

In the following, we introduce the complete VECTOR pipeline for the three-way allocation. The pipeline is designed as a lightweight, plug-and-play extension that augments 
% \jt{remove "any", any is too strong} any 
existing token-importance-based eviction \rev{algorithms} (e.g., SnapKV~\citep{li2024snapkv}, KVzip~\citep{kim2025kvzip}, KeyDiff~\cite{park2025keydiff}).

Denote $p_c$ as the target memory compression ratio. To incorporate the \textit{Approximation} state while preserving the same memory budget, VECTOR executes the following three-step pipeline: 
% \jt{we have figure 2, but we never refer to it in the main content. The figure is to help us illustate instead of just putting it there}

\textbf{Step 1: Budget Relaxation.} Let $p_a$ denote the approximation ratio. VECTOR invokes the base token-importance eviction algorithm to identify an expanded candidate pool of size $1-p_c+p_a$. This yields a larger initial set of contextually relevant tokens, from which we subsequently determine which ones to compress.

\textbf{Step 2: Residual Evaluation.} For each token in the expanded candidate pool, VECTOR computes the per-token reconstruction error $\varepsilon_i = \| V_i - W_\textrm{OLS} K_i \|^2_2$ using the offline-calibrated OLS projection matrix $W_\textrm{OLS}$. This step introduces an orthogonal dimension of evaluation---\textit{reconstructability}---that quantifies the approximation error incurred if the token's value were discarded.

\textbf{Step 3: Asymmetric Truncation.} As illustrated in Figure~\ref{fig:orthogonal}, VECTOR partitions the expanded pool based on the computed reconstruction errors to execute the budget swap:

\begin{itemize}[leftmargin=*,itemsep=1pt,topsep=3pt]
    \item The keys of \textit{all} $1-p_c+p_a$ tokens in the expanded pool are retained.
    \item The $2p_a$ tokens with the lowest reconstruction errors $\varepsilon_i$ (i.e., those most amenable to approximation) have their cached values discarded. \rev{Their values are reconstructed from the retained keys as $\hat{V} = W_\textrm{OLS} K$.}
    % [REV-DEL] During generation, these values are dynamically reconstructed via $\hat{V} = W_\textrm{OLS} K$.
    \item Both keys and values are exactly retained for the remaining $1-p_c-p_a$ proportion of tokens.
\end{itemize}

Through the above procedure, $1-p_c+p_a$ keys and $1-p_c-p_a$ values are stored in memory, consuming the exact same footprint as $1-p_c$ full KV pairs. Specifically, the \textit{Eviction}, \textit{Approximation}, and \textit{Retention} states account for proportions $p_c-p_a$, $2p_a$, and $1-p_c-p_a$ of all tokens, respectively. 
% The choice of $p_a$ is discussed in Section~\ref{sec:theory}, along with a sensitivity analysis in Appendix~\ref{app:pa_sensitivity}.
% [REV-DEL] The choice of $p_a$ is discussed in Section~\ref{sec:theory}, along with a sensitivity analysis in Section~\ref{sec:experiments:pa_sensitivity}.

\rev{\paragraph{Memory Budget} The compression ratio $p_c$ fixes the total memory budget, and $p_a$ only decides how this budget is spent. Retaining a token costs a key and a value, approximating it costs only a key, and evicting it costs nothing. Increasing $p_a$ by $\delta$ moves $\delta$ tokens from retention and $\delta$ tokens from eviction into approximation, so the total memory stays unchanged. As a result, VECTOR supports any compression ratio that the base eviction method supports, and setting $p_a=0$ recovers the base method. Under the same budget, the approximation tier lets $1-p_c+p_a$ tokens keep their keys instead of $1-p_c$. As in the base eviction methods, all routing decisions are made once at the end of prefill (Appendix~\ref{app:systems}).}

{\color{\revcolor}
\paragraph{When Does Approximation Help?} Let $w_i$ be the importance score of token $i$, normalized to sum to one. We measure the distortion after compression by $\mathcal{E}$, the importance-weighted signal loss. An evicted token contributes $w_i$, a retained token contributes $0$, and an approximated token contributes $w_i$ times its normalized reconstruction error $\|V_i-\hat{V}_i\|_2^2/\|V_i\|_2^2$. Let $R^2_\textrm{approx}$ be the OLS prediction quality over the approximated tokens, $w^*$ the importance score at the eviction boundary, and $\bar{w}$ the average importance score in the expanded pool.

\begin{proposition}\label{prop:skewness_threshold}
Assume that the per-token reconstruction errors are independent of the importance scores, and that $p_a$ is small. Increasing $p_a$ reduces $\mathcal{E}$ if
\begin{equation*}
    R^2_\textrm{approx} > \frac{\bar{w}}{w^*+\bar{w}}.
\end{equation*}
\end{proposition}

\noindent The proof is given in Appendix~\ref{sec:appendix:proof:prop}. The approximation tier therefore helps most when the evicted tokens still carry substantial importance, that is, when $w^*$ is large relative to $\bar{w}$.
}

{\color{\revcolor}
\paragraph{Choice of $p_a$} Since $w^*$ and $\bar{w}$ vary across samples, we set $p_a$ with a fixed rule instead of tuning it per sample:
\begin{equation} \label{eq:p_a_deploy}
    p_a^{\mathrm{deploy}} = \min\!\left(p_c/2,\; (1-p_c-\epsilon)/2\right),
\end{equation}
where $\epsilon>0$ is a small constant ensuring a non-empty \textit{Retention}
tier. The first term keeps the \textit{Approximation} tier no larger than half
the compressed budget, and the second ensures sufficient headroom for the
\textit{Retention} tier. A sensitivity analysis of $p_a$ is given in Section~\ref{sec:experiments:pa_sensitivity}.
}

% Section \ref{sec:theory} details the formal theoretical formulation, including how to optimally determine the approximation ratio and formulate the rate-distortion bounds.

% [REV] Theory moved to the appendix (see sections/appendix.tex)
% \input{sections/theory}
\section{Experiments}\label{sec:experiments}

% \yue{In the following, we present the main experiments results. Due to page limit, we postpone ablation studies in the appendix for the comparison between OLS and the MP pseudo-inverse mentioned in Section \ref{sec:method} (Section \ref{app:ols_vs_mp}), the comparison between V-only and K-only approximation (Section \ref{app:vonly_vs_konly}), and the algorithm's sensitivity to the approximation ratio $p_a$ (Section \ref{app:pa_sensitivity}). }
% We present the main experimental results below. Due to page limitations, ablation studies are deferred to the appendix. These include a comparison between OLS and the analytical pseudo-inverse (Appendix~\ref{app:ols_vs_mp}), a comparison between V-only and K-only approximation (Appendix~\ref{app:vonly_vs_konly}), and a sensitivity analysis of the approximation ratio $p_a$ (Appendix~\ref{app:pa_sensitivity}).
We present the main experimental results below. Due to page limitations, some \rev{results} are deferred to the appendix. These include a comparison between OLS and the analytical pseudo-inverse (Appendix~\ref{app:ols_vs_mp})\rev{,} a comparison between V-only and K-only approximation (Appendix~\ref{app:vonly_vs_konly})\rev{, a robustness study of OLS calibration (Appendix~\ref{app:calib}), and a multi-turn evaluation on SCBench (Appendix~\ref{app:scbench})}.

% \begin{itemize}
%     \item Experimental setup
%     \item Benchmark main results
%     \item NIAH results
%     \item Discussion
%     \begin{itemize}
%         \item Improvement under high compression rate
%         \item Generalization across tasks, across models, across baseline methods
%     \end{itemize}
% \end{itemize}

% [REV] Per-task LongBench table moved to the appendix (tab:exp_main_full); replaced by the summary table below.
\begin{table*}[t]
\centering
\small
{\color{\revcolor}
\setlength{\tabcolsep}{3.2pt}
\resizebox{\textwidth}{!}{%
\begin{tabular}{lcccccccccccc}
\toprule
\multirow{2}{*}{Method} & \multicolumn{4}{c}{\textbf{Llama-3.1-8B}} & \multicolumn{4}{c}{\textbf{Qwen3-14B}} & \multicolumn{4}{c}{\textbf{Qwen3-0.6B}} \\
\cmidrule(lr){2-5}\cmidrule(lr){6-9}\cmidrule(lr){10-13}
 & $p_c{=}0.25$ & $p_c{=}0.50$ & $p_c{=}0.75$ & $p_c{=}0.90$ & $p_c{=}0.25$ & $p_c{=}0.50$ & $p_c{=}0.75$ & $p_c{=}0.90$ & $p_c{=}0.25$ & $p_c{=}0.50$ & $p_c{=}0.75$ & $p_c{=}0.90$ \\
\midrule
No Comp. & \multicolumn{4}{c}{45.69} & \multicolumn{4}{c}{50.08} & \multicolumn{4}{c}{30.93} \\
\midrule
KeyDiff & 47.70 & 46.15 & 41.65 & 35.28 & 46.61 & 40.72 & 32.23 & 22.71 & 28.06 & 23.13 & 16.72 & 13.36 \\
\quad + VECTOR & 46.16\,{\scriptsize\textcolor{gray}{-1.5}} & \textbf{47.68}\,{\scriptsize\textcolor{teal}{+1.5}} & \textbf{44.86}\,{\scriptsize\textcolor{teal}{+3.2}} & \textbf{38.75}\,{\scriptsize\textcolor{teal}{+3.5}} & \textbf{49.48}\,{\scriptsize\textcolor{teal}{+2.9}} & \textbf{47.75}\,{\scriptsize\textcolor{teal}{+7.0}} & \textbf{41.38}\,{\scriptsize\textcolor{teal}{+9.2}} & \textbf{32.44}\,{\scriptsize\textcolor{teal}{+9.7}} & \textbf{30.25}\,{\scriptsize\textcolor{teal}{+2.2}} & \textbf{28.91}\,{\scriptsize\textcolor{teal}{+5.8}} & \textbf{22.88}\,{\scriptsize\textcolor{teal}{+6.2}} & \textbf{18.94}\,{\scriptsize\textcolor{teal}{+5.6}} \\
\midrule
SnapKV$^\dagger$ & 45.87 & 45.89 & 45.51 & 43.38 & 50.06 & 49.81 & 49.13 & 46.29 & 30.80 & 31.18 & 31.05 & 29.85 \\
\quad + VECTOR$^\dagger$ & 45.46\,{\scriptsize\textcolor{gray}{-0.4}} & 45.08\,{\scriptsize\textcolor{gray}{-0.8}} & 45.11\,{\scriptsize\textcolor{gray}{-0.4}} & \textbf{45.20}\,{\scriptsize\textcolor{teal}{+1.8}} & 49.79\,{\scriptsize\textcolor{gray}{-0.3}} & 49.01\,{\scriptsize\textcolor{gray}{-0.8}} & 48.63\,{\scriptsize\textcolor{gray}{-0.5}} & \textbf{47.65}\,{\scriptsize\textcolor{teal}{+1.4}} & \textbf{30.99}\,{\scriptsize\textcolor{teal}{+0.2}} & 30.47\,{\scriptsize\textcolor{gray}{-0.7}} & 30.96\,{\scriptsize\textcolor{gray}{-0.1}} & \textbf{31.48}\,{\scriptsize\textcolor{teal}{+1.6}} \\
\midrule
KVzip & 45.23 & 43.82 & 44.95 & 41.23 & 49.90 & 49.98 & 49.49 & 44.75 & 30.83 & 31.39 & 29.30 & 24.28 \\
\quad + VECTOR & \textbf{45.49}\,{\scriptsize\textcolor{teal}{+0.3}} & \textbf{45.08}\,{\scriptsize\textcolor{teal}{+1.3}} & \textbf{45.11}\,{\scriptsize\textcolor{teal}{+0.2}} & \textbf{45.20}\,{\scriptsize\textcolor{teal}{+4.0}} & 49.58\,{\scriptsize\textcolor{gray}{-0.3}} & 48.82\,{\scriptsize\textcolor{gray}{-1.2}} & 49.14\,{\scriptsize\textcolor{gray}{-0.4}} & \textbf{47.53}\,{\scriptsize\textcolor{teal}{+2.8}} & \textbf{30.95}\,{\scriptsize\textcolor{teal}{+0.1}} & 29.96\,{\scriptsize\textcolor{gray}{-1.4}} & \textbf{31.45}\,{\scriptsize\textcolor{teal}{+2.1}} & \textbf{27.05}\,{\scriptsize\textcolor{teal}{+2.8}} \\
\midrule
PyramidKV$^\dagger$ & 47.56 & 46.97 & 46.00 & 43.52 & 50.83 & 49.82 & 48.60 & 45.89 & 26.24 & 25.46 & 27.14 & 28.29 \\
\quad + VECTOR$^\dagger$ & \textbf{47.63}\,{\scriptsize\textcolor{teal}{+0.1}} & \textbf{47.31}\,{\scriptsize\textcolor{teal}{+0.3}} & \textbf{46.54}\,{\scriptsize\textcolor{teal}{+0.5}} & \textbf{45.06}\,{\scriptsize\textcolor{teal}{+1.5}} & 50.69\,{\scriptsize\textcolor{gray}{-0.1}} & 49.29\,{\scriptsize\textcolor{gray}{-0.5}} & \textbf{48.70}\,{\scriptsize\textcolor{teal}{+0.1}} & 45.56\,{\scriptsize\textcolor{gray}{-0.3}} & \textbf{26.31}\,{\scriptsize\textcolor{teal}{+0.1}} & \textbf{26.27}\,{\scriptsize\textcolor{teal}{+0.8}} & 25.29\,{\scriptsize\textcolor{gray}{-1.9}} & 24.30\,{\scriptsize\textcolor{gray}{-4.0}} \\
\bottomrule
\end{tabular}}
\caption{LongBench results (16-task average) for four eviction baselines and their VECTOR-augmented variants on three models, under compression ratios $p_c \in \{0.25,0.50,0.75,0.90\}$. Small numbers give the change over the base method, and bold marks improvements. Approximation ratios are set by Eq.~\eqref{eq:p_a_deploy}. $^\dagger$ denotes query-aware compression. Per-task results are in Appendix~\ref{app:additional_results}.}
\label{tab:exp_main}
\vspace{-1.0em}% [REV-VSPACE]
}
\end{table*}

\subsection{Experimental Setup}\label{sec:experiments:setup}
We conduct all experiments with the KVPress evaluation framework~\citep{devoto2025expected} under a unified protocol for long-context compression.

% Before downstream evaluation, VECTOR performs a one-time offline calibration to fit the layer-wise linear map $W_\mathrm{OLS}$ used for value reconstruction. Following the protocol in Section~\ref{sec:approximation}, we use C4 dataset~\citep{dodge2021documenting} to collect key--value activation pairs from 10,000 sequences (length 4,096) per model, fit OLS per transformer layer, and evaluate predictability on a held-out set of 100 sequences (length 4,096). The resulting calibrated weights are then fixed for all LongBench and NIAH experiments and are not updated during inference.
Before downstream evaluation, VECTOR performs a one-time offline calibration to fit the layer-wise linear map $W_\mathrm{OLS}$ used for value reconstruction. Following the protocol in Section~\ref{sec:approximation}, we use C4 dataset~\citep{dodge2021documenting} to collect key--value activation pairs from \rev{5,000 sequences (Llama-3.1-8B) or 10,000 sequences (Qwen3 models) of 2,048 tokens}, and fit OLS per transformer layer. The resulting calibrated weights are then fixed for all \rev{downstream} experiments.
% [REV-DEL] 10,000 sequences (length 4,096) per model ... fixed for all LongBench and NIAH experiments.

% \paragraph{Baselines} We evaluate four representative eviction baselines and their VECTOR-augmented variants: KeyDiff~\citep{park2025keydiff}, SnapKV~\citep{li2024snapkv}, KVzip~\citep{kim2025kvzip}, and PyramidKV~\citep{cai2024pyramidkv}. For each baseline, the corresponding ``+ VECTOR'' method keeps the original importance scorer and adds the reconstructability-aware approximation tier. \yp{mention that snapkv and pyramidkv is query-aware}
\paragraph{Baselines}
We evaluate four representative token-eviction baselines and their VECTOR-augmented variants: KeyDiff~\citep{park2025keydiff}, SnapKV~\citep{li2024snapkv}, KVzip~\citep{kim2025kvzip}, and PyramidKV~\citep{cai2024pyramidkv}. These methods differ in how they score token importance. \textit{Query-aware} methods (SnapKV, PyramidKV) identify important tokens via the attention patterns that the query induces over the context, producing task-relevant eviction decisions that are tailored to the specific question being asked. \textit{Query-agnostic} methods (KeyDiff, KVzip) score tokens using properties intrinsic to the context itself (i.e., key geometric diversity for KeyDiff and semantic redundancy for KVzip) without access to the query. For each baseline, the corresponding ``+\,VECTOR'' variant retains the original importance scorer and augments it with the reconstructability-aware approximation tier.

% \paragraph{Models} We report results on three open-source backbones: \texttt{Llama-3.1-8B-Instruct}~\citep{grattafiori2024llama}, \texttt{Qwen3-14B}~\citep{yang2025qwen3}, and \texttt{Qwen3-0.6B}~\citep{yang2025qwen3}.
\rev{\paragraph{Competing Methods}
We further compare with six compression methods: ChunkKV~\citep{liu2025chunkkv}, ExpectedAttention~\citep{devoto2025expected}, AdaKV-SnapKV~\citep{feng2024adakv}, TOVA~\citep{oren2024tova}, D2O~\citep{wan2024d2o}, and RocketKV~\citep{behnam2025rocketkv}. All methods use the same prompts, scorer, and compression ratios as the main experiments. Protocol details are given in Appendix~\ref{app:competing}.}

\paragraph{Models} 
We report results on three open-source models: \texttt{Llama-3.1-8B-Instruct}~\citep{grattafiori2024llama}, \texttt{Qwen3-14B}~\citep{yang2025qwen3}, and \texttt{Qwen3-0.6B}~\citep{yang2025qwen3}.

% \paragraph{Benchmark} We use LongBench~\citep{bai2024longbench} as the primary quantitative benchmark and Needle-in-a-Haystack (NIAH)~\citep{kamradt2023niah} as a stress test for long-range retrieval. For LongBench, we evaluate on a 16-task setting by excluding the 5 Chinese language tasks from the original 21-task collection, and evaluate on the full test split.
\paragraph{Benchmarks}
We use LongBench~\citep{bai2024longbench} as the primary quantitative benchmark and Needle-in-a-Haystack (NIAH)~\citep{kamradt2023niah} as a stress test for long-range retrieval~\rev{(Appendix~\ref{app:niah})}. For LongBench, we evaluate on a 16-task subset obtained by excluding the 5 Chinese-language tasks from the original 21-task collection, using the full test split. \rev{We further evaluate long-context retrieval with RULER~\citep{hsieh2024ruler} at context lengths from 16K to 128K tokens, and multi-turn reuse of a compressed cache with SCBench~\citep{li2025scbench} in Appendix~\ref{app:scbench}.}

% \paragraph{Metrics} LongBench is evaluated with task-specific official metrics (e.g., F1, ROUGE-L, accuracy/EM, and edit similarity), and we report \texttt{Avg} as the arithmetic mean over the 16 selected tasks. For compression, we evaluate $p_c \in \{0.25, 0.50, 0.75, 0.90\}$ and set approximation ratios by Eq.~\ref{eq:p_a_deploy}: $p_a=\{0.125, 0.20, 0.125, 0.05\}$, respectively.
\paragraph{Metrics}
LongBench tasks are evaluated with their official task-specific metrics (F1, ROUGE-L, exact match / accuracy, and edit similarity); we report \texttt{Avg} as the arithmetic mean over the 16 selected tasks. We evaluate compression ratios $p_c \in \{0.25, 0.50, 0.75, 0.90\}$, with the corresponding approximation ratios set by Eq.~\eqref{eq:p_a_deploy}: $p_a \in \{0.125, 0.25, 0.125, 0.05\}$, respectively. \rev{Since greedy decoding is deterministic, we quantify uncertainty with a paired bootstrap over test samples~\citep{koehn2004statistical} with 10,000 resamples (Appendix~\ref{app:significance}).}

More implementation and reproducibility details are provided in Appendix~\ref{app:exp_repro}.

\subsection{LongBench Main Results}

% Table~\ref{tab:exp_main} reports full LongBench results. Overall, augmenting eviction baselines with VECTOR improves the quality-memory trade-off in the majority of evaluated settings, with gains typically increasing alongside the compression ratio, a regime where binary eviction is more susceptible to irreversible information loss.
% [REV-DEL] {Table~\ref{tab:exp_main} reports full LongBench results. Overall, augmenting eviction baselines with VECTOR improves downstream performance in most high-compression settings, with gains most consistent at $p_c \in \{0.75, 0.90\}$. At lower compression ratios, the approximation tier provides limited additional benefit, as the base eviction policy already retains most contextually relevant tokens.}
\rev{Table~\ref{tab:exp_main} reports the LongBench results. Augmenting eviction baselines with VECTOR improves downstream performance under high compression, where binary eviction discards the most information. At $p_c=0.90$, VECTOR improves 10 of the 12 base methods across the three models.}

The strongest and most consistent improvements are observed with the two query-agnostic baselines, KeyDiff and KVzip. For KeyDiff, gains are particularly pronounced at high compression: on Qwen3-14B, KeyDiff+VECTOR improves the average score by $+7.03$ points at $p_c{=}0.50$, $+9.15$ at $p_c{=}0.75$, and $+9.73$ at $p_c{=}0.90$\rev{.}
% on Qwen3-0.6B, the corresponding gains are $+5.78$, $+6.16$, and $+5.58$ points, respectively. 
Similar gains are observed on Qwen3-0.6B~\rev{(Table~\ref{tab:exp_main})}.
% [REV-DEL] KVzip+VECTOR delivers consistent improvements at high compression across all three models, reaching $+3.97$ points on Llama-3.1-8B at $p_c{=}0.90$.
\rev{On Llama-3.1-8B, the gains of KeyDiff+VECTOR are significant at both $p_c=0.75$ and $p_c=0.90$ (Appendix~\ref{app:significance}). KVzip+VECTOR improves over KVzip at $p_c=0.90$ on all three models, by $+3.97$, $+2.78$, and $+2.77$ points on Llama-3.1-8B, Qwen3-14B, and Qwen3-0.6B.} These results support the central claim of this work: introducing a reconstructability-aware approximation tier can recover a substantial portion of the information that would otherwise be irretrievably lost under binary eviction.
\rev{For the query-aware baselines, VECTOR is most effective at the highest compression ratio. At $p_c=0.90$, SnapKV+VECTOR improves over SnapKV on all three models, by $+1.82$, $+1.36$, and $+1.63$ points, and the gain on Llama-3.1-8B is significant (95\% CI $[+1.06, +2.54]$). As shown in Section~\ref{sec:method_pipeline}, the approximation tier helps most when the evicted tokens still carry substantial importance. This holds under high compression, and also with query-agnostic scorers, which select tokens without seeing the question.}

\rev{\subsection{Comparison with Competing Methods}\label{sec:experiments:competing}}

\begin{table}[t]
\centering
\small
{\color{\revcolor}
\setlength{\tabcolsep}{2.5pt}
\begin{tabular}{lcccc}
\toprule
\multirow{2}{*}{Method} & \multicolumn{4}{c}{Compression ratio $p_c$} \\
\cmidrule(lr){2-5}
 & 0.25 & 0.50 & 0.75 & 0.90 \\
\midrule
No Comp. & \multicolumn{4}{c}{45.69} \\
\midrule
ChunkKV$^\dagger$ & 45.60 & 45.08 & 43.94 & 40.65 \\
ExpectedAttention & 46.48 & 47.19 & 43.59 & 33.57 \\
AdaKV-SnapKV$^\dagger$ & 45.72 & 45.77 & 44.85 & 43.54 \\
TOVA$^\dagger$ & 45.29 & 44.68 & 43.83 & 41.62 \\
D2O & 45.27 & 44.92 & 44.23 & 43.25 \\
RocketKV & 45.88 & 45.28 & 44.34 & 42.90 \\
\midrule
KeyDiff + VECTOR & 46.16 & \textbf{47.68} & 44.86 & 38.75 \\
SnapKV + VECTOR$^\dagger$ & 45.46 & 45.08 & 45.11 & \textbf{45.20} \\
KVzip + VECTOR & 45.49 & 45.08 & 45.11 & \textbf{45.20} \\
PyramidKV + VECTOR$^\dagger$ & \textbf{47.63} & 47.31 & \textbf{46.54} & 45.06 \\
\bottomrule
\end{tabular}
\caption{LongBench average on Llama-3.1-8B for six competing compression methods and the four VECTOR-augmented methods, under compression ratios $p_c \in \{0.25, 0.50, 0.75, 0.90\}$. Bold marks the best method at each compression ratio. $^\dagger$ denotes query-aware compression. Results of the base eviction methods without VECTOR are in Table~\ref{tab:exp_main}. Results for Qwen3-14B and Qwen3-0.6B are in Appendix~\ref{app:competing}.}
\label{tab:exp_competing}
\vspace{-1.0em}% [REV-VSPACE]
}
\end{table}

\rev{Table~\ref{tab:exp_competing} compares the VECTOR-augmented methods with six competing compression methods on Llama-3.1-8B. From $p_c=0.50$ upward, the best method is a VECTOR-augmented one. At $p_c=0.90$, SnapKV+VECTOR and KVzip+VECTOR reach 45.20, which is 1.66 points above the strongest competitor, AdaKV-SnapKV, and 98.9\% of the uncompressed score. SnapKV+VECTOR also outperforms all competing methods at $p_c=0.90$ on Qwen3-14B (47.65 vs.\ 47.18) and Qwen3-0.6B (31.48 vs.\ 29.84; Appendix~\ref{app:competing}). Note that VECTOR is a plug-in for eviction methods rather than a standalone compression method. Several competing methods, such as ChunkKV, AdaKV-SnapKV, and TOVA, are themselves importance-based eviction policies and could be augmented with VECTOR in the same way.}

\rev{\subsection{Long-Context Evaluation on RULER}\label{sec:experiments:ruler}}

\begin{table}[t]
\centering
\small
{\color{\revcolor}
\setlength{\tabcolsep}{2.5pt}
\begin{tabular}{lcccc}
\toprule
\multirow{2}{*}{Method} & \multicolumn{4}{c}{Context length} \\
\cmidrule(lr){2-5}
 & 16K & 32K & 64K & 128K \\
\midrule
No Comp. & 94.48 & 88.30 & 85.66 & 80.16 \\
\midrule
KeyDiff & 65.65 & \textbf{64.91} & \textbf{62.62} & 55.67 \\
KeyDiff + VECTOR & \textbf{67.17} & 64.27 & 61.28 & \textbf{57.19} \\
\midrule
SnapKV$^\dagger$ & 69.35 & 72.74 & 73.68 & 69.12 \\
SnapKV + VECTOR$^\dagger$ & \textbf{82.11} & \textbf{78.29} & \textbf{76.32} & \textbf{70.69} \\
\midrule
KVzip & 82.61 & 75.95 & 67.26 & 45.42 \\
KVzip + VECTOR & \textbf{88.84} & \textbf{83.59} & \textbf{75.57} & \textbf{55.97} \\
\midrule
PyramidKV$^\dagger$ & 68.77 & 70.97 & \textbf{73.84} & 67.41 \\
PyramidKV + VECTOR$^\dagger$ & \textbf{75.68} & \textbf{76.01} & 73.80 & \textbf{69.40} \\
\bottomrule
\end{tabular}
\caption{RULER score (mean over 13 subtasks) on Llama-3.1-8B at $p_c=0.90$ for context lengths from 16K to 128K tokens. Bold marks the better of each base method and its VECTOR variant. Results at $p_c=0.75$ are in Appendix~\ref{app:ruler}. $^\dagger$ denotes query-aware compression.}
\label{tab:exp_ruler}
\vspace{-1.0em}% [REV-VSPACE]
}
\end{table}

\rev{To evaluate VECTOR at long context, we run RULER~\citep{hsieh2024ruler} on Llama-3.1-8B with contexts from 16K to 128K tokens, using the same methods and settings as above. Table~\ref{tab:exp_ruler} reports results at $p_c=0.90$. Across the four base methods and four context lengths, VECTOR improves 13 of the 16 settings, and 9 of these gains are significant, with no significant drop (Appendix~\ref{app:significance}). The gains hold up to 128K tokens. KVzip drops to 45.42 at 128K, and KVzip+VECTOR raises it to 55.97 ($+10.55$). At 16K, SnapKV+VECTOR improves SnapKV by 12.76 points. Since VECTOR only rewrites the KV cache after prefill, it runs with standard fused attention up to 128K tokens on a single GPU. NIAH heatmaps at $p_c=0.90$ show the same pattern (Appendix~\ref{app:niah}).}

\rev{\subsection{Reconstructability Signal}\label{sec:experiments:signal}
VECTOR selects the approximated tokens by reconstruction error. To test whether this signal matters, we keep the candidate pool and the budget fixed, and replace the selection rule with importance ranking, which approximates the least important candidates. We use KeyDiff+VECTOR on Llama-3.1-8B with six LongBench QA tasks at $p_c \in \{0.75, 0.90\}$ (Appendix~\ref{app:signal_setup}). Selecting by reconstruction error outperforms selecting by importance by 2.31 points on average, and wins in 10 of the 12 settings. The correlation between importance and reconstruction error is only 0.16. The two signals are nearly independent, so importance cannot replace the reconstruction error when choosing which tokens to approximate.}

\rev{\subsection{Inference Overhead}\label{sec:experiments:overhead}
We measure decoding throughput and prefill time at a 32K context and $p_c=0.90$, for KeyDiff, SnapKV, and PyramidKV on all three models (Appendix~\ref{app:systems}). In all nine settings, VECTOR and its base method have the same decoding throughput within measurement noise, because both decode with standard attention over a compressed cache. Value reconstruction adds a one-time prefill cost of 0.16 to 0.21 seconds, which is 6\% of the prefill time for Qwen3-14B and 9\% for Llama-3.1-8B. The OLS matrices take 67\,MB for Llama-3.1-8B, independent of context length and batch size.}
%%%%%%%%%%%%%%%%%%%% [REV] end of new subsections %%%%%%%%%%%%%%%%%%%%

% [REV] NIAH figure moved to the appendix (app:niah)

\subsection{Sensitivity to Approximation Ratio}
\label{sec:experiments:pa_sensitivity}

\begin{figure}[t]
    \centering
    \includegraphics[width=0.9\columnwidth]{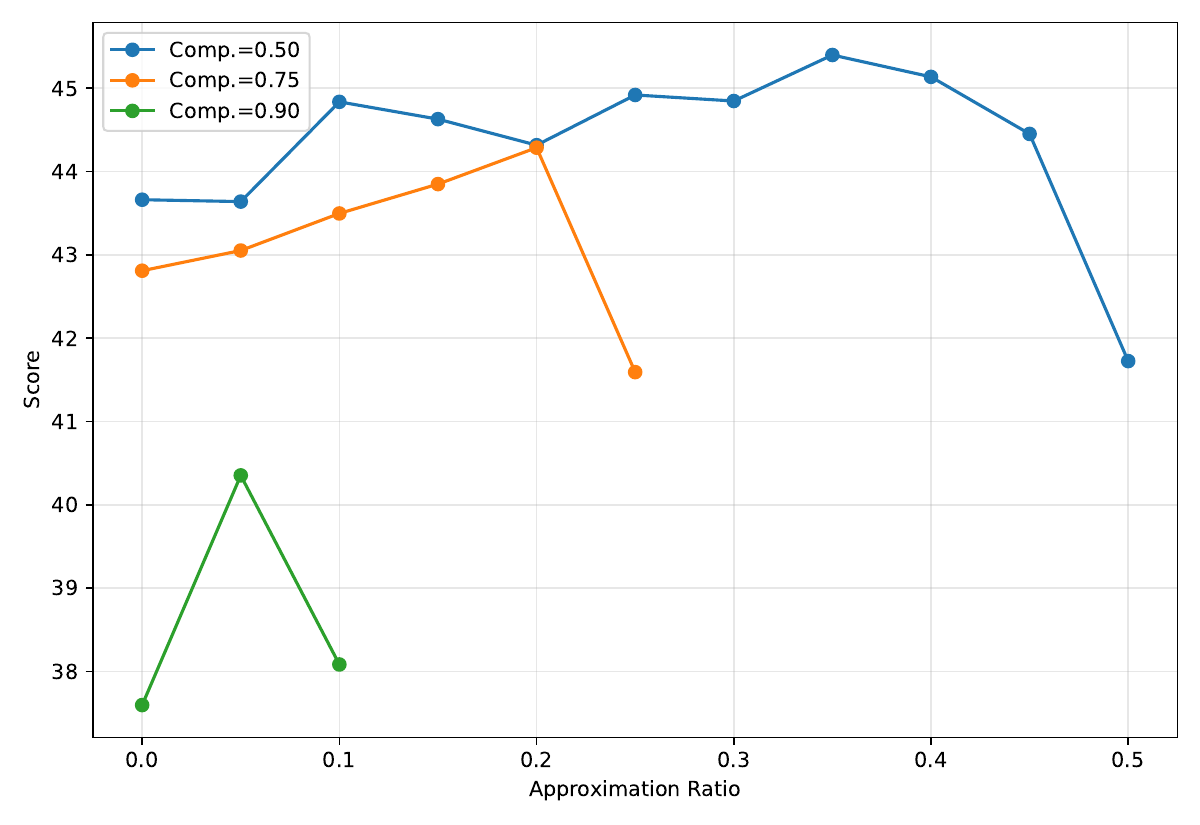}
    \caption{Mean LongBench score vs.\ approximation ratio $p_a$ under three compression ratios, averaged over two baselines (KeyDiff, KVzip) and two tasks (HotpotQA, NarrativeQA) on Llama-3.1-8B-Instruct.}
    \label{fig:pa_sensitivity}
\vspace{-1.0em}% [REV-VSPACE]
\end{figure}

\rev{Eq.~\eqref{eq:p_a_deploy} sets $p_a$ with a fixed rule. To test how sensitive VECTOR is to this choice, we sweep $p_a$ at $p_c \in \{0.50, 0.75, 0.90\}$ with two base methods on two LongBench tasks (setup in Appendix~\ref{app:pa_sweep}). As shown in Figure~\ref{fig:pa_sensitivity}, performance first rises and then falls as $p_a$ grows at all three compression ratios. The values given by Eq.~\eqref{eq:p_a_deploy} ($p_a=0.25$, $0.125$, and $0.05$) all fall in the region where performance improves over binary eviction ($p_a=0$), and at $p_c=0.90$ the rule matches the peak. A small $p_a$ leaves recoverable tokens evicted. A large $p_a$ shrinks the Retention tier, so hard-to-reconstruct tokens are approximated, and it spends budget on low-importance tokens that would be better evicted.}
\section{Conclusion}\label{sec:conclusion}

% This paper studies KV cache compression from a reconstructability-aware perspective and proposes VECTOR, a lightweight three-way allocation framework that can be integrated with existing eviction methods. 
We propose VECTOR, a lightweight three-way allocation framework that integrates reconstructability-aware approximation into existing eviction methods.
By separating token importance from reconstructability, VECTOR extends binary retain/evict decisions into retention, approximation, and eviction under the same memory budget. % [REV-DEL] The design is practical: value approximation is implemented with offline OLS calibration and does not require model retraining.
%
% \yue{
% [REV-DEL] Theoretical discussions are provided to connect the OLS $R^2$ with the compression performance.
\rev{Value approximation uses an offline OLS calibration without model retraining, and our analysis links the OLS $R^2$ to the reduction in distortion.}
% } 
% [REV-DEL] Empirically, augmenting representative baselines with VECTOR improves long-context performance in most high-compression settings and yields more robust retrieval behavior in NIAH.
\rev{Empirically, augmenting representative baselines with VECTOR yields significant gains under high compression on LongBench, RULER with up to 128K tokens, and multi-turn SCBench. At $p_c=0.90$, the best VECTOR-augmented method outperforms six competing compression methods, and VECTOR adds no measurable decoding overhead.}
 \rev{These results suggest that combining importance-aware eviction with reconstructability-aware approximation is a promising direction for long-context inference.}
% [REV-DEL] These results suggest that combining importance-aware eviction with reconstructability-aware approximation is a promising direction for future long-context inference systems.

\section*{Limitations}
% [REV-DEL] VECTOR has some limitations. First, $p_a$ is set by an empirical formula (Eq.~\eqref{eq:p_a_deploy}) rather than optimized per-sample or per-layer. This may leave room for improvement, particularly for tasks or models where the optimal $p_a$ deviates from the formula. Second, gains over query-aware baselines such as SnapKV and PyramidKV are modest. These methods already preserve query-relevant tokens effectively, leaving limited headroom for the approximation tier. To improve, scorer-aware allocation and adaptive approximation ratios can be potential future directions.
\rev{VECTOR has some limitations. First, $p_a$ is set by a fixed rule (Eq.~\eqref{eq:p_a_deploy}) that is shared by all base scorers, budgets, and layers. Adapting $p_a$ to the base scorer and the budget is a natural extension. For example, query-aware scorers at moderate budgets already retain most query-relevant tokens and may benefit from a smaller approximation tier (Appendix~\ref{app:ruler}). Second, our implementation is built on KVPress without custom kernels. A fused attention kernel that stores only keys for approximated tokens would realize the memory layout of Section~\ref{sec:method_pipeline} directly (Appendix~\ref{app:systems}).}

\bibliography{ref}

\appendix

\section{Additional Results}
\label{app:additional_results}

\rev{Tables~\ref{tab:exp_main_full} and~\ref{tab:exp_qwen06b} report the per-task LongBench results summarized in Table~\ref{tab:exp_main}.}

% [REV] Moved from the main text
\begin{table*}[p]
  \centering
  \resizebox{\textwidth}{!}{
  \begin{tabular}{lcccccccccccccccccc}
    \toprule
    & & \multicolumn{16}{c}{LongBench} & \\
    \cmidrule(r){3-18}
    Method & \makecell{Comp. \\Ratio} & narrativeqa & qasper & \makecell{multifield- \\ qa\_en} & hotpotqa & 2wikimqa & musique & gov\_report & qmsum & multi\_news & trec & triviaqa & samsum & \makecell{passage\_ \\ count} & \makecell{passage\_ret- \\ rieval\_en} & lcc & repobench-p & Avg. \\
    
    \midrule
    
    \textbf{Llama-3.1-8B} & No Comp. & 30.59 & 46.93 & 55.94 & 59.04 & 51.99 & 33.17 & 35.22 & 25.19 & 26.84 & 29.00 & 91.56 & 40.11 & 10.15 & 100.00 & 51.28 & 44.11 & 45.69 \\
    \cmidrule(r){1-19}
     \multirow{4}{5em}{KeyDiff} & 0.25 & 31.27 & 47.30 & 53.98 & 58.39 & 50.69 & 34.08 & 34.54 & 24.51 & 26.80 & 63.00 & 92.71 & 41.46 & 11.15 & 100.00 & 49.04 & 44.23 & \textbf{47.70} \\
     & 0.50 & 32.46 & 45.65 & 52.60 & 56.07 & 46.20 & 28.62 & 32.98 & 24.33 & 25.79 & 64.00 & 92.71 & 41.28 & 10.15 & 99.50 & 42.65 & 43.48 & 46.15 \\
     & 0.75 & 29.14 & 31.75 & 43.25 & 52.83 & 38.74 & 24.69 & 29.36 & 23.37 & 22.99 & 49.50 & 91.93 & 40.01 & 10.56 & 99.50 & 34.33 & 44.44 & 41.65 \\
     & 0.90 & 27.53 & 17.00 & 34.69 & 40.03 & 26.18 & 16.01 & 26.12 & 21.28 & 18.89 & 40.00 & 92.24 & 40.30 & 9.50 & 89.00 & 21.33 & 44.34 & 35.28 \\
    \cmidrule(r){1-19}
     \multirow{4}{5em}{KeyDiff\\ + VECTOR} & 0.25 & 30.60 & 48.18 & 56.07 & 58.45 & 50.72 & 33.59 & 34.80 & 24.74 & 26.90 & 34.50 & 92.71 & 41.17 & 10.65 & 99.50 & 51.63 & 44.29 & 46.16 \\
     & 0.50 & 32.24 & 47.38 & 56.77 & 58.92 & 51.65 & 35.47 & 32.99 & 24.68 & 26.10 & 59.00 & 92.49 & 40.54 & 10.63 & 99.50 & 50.73 & 43.78 & \textbf{47.68} \\
     & 0.75 & 30.54 & 39.20 & 53.15 & 54.83 & 45.85 & 28.83 & 29.55 & 23.79 & 24.03 & 56.50 & 91.87 & 39.67 & 10.58 & 98.50 & 46.56 & 44.26 & \textbf{44.86} \\
     & 0.90 & 28.12 & 23.96 & 37.98 & 49.88 & 30.72 & 20.72 & 25.75 & 22.39 & 20.43 & 43.00 & 91.19 & 39.10 & 12.05 & 94.50 & 35.20 & 45.03 & \textbf{38.75} \\
    \cmidrule(r){1-19}
     \multirow{4}{5em}{SnapKV} & 0.25 & 30.59 & 46.85 & 56.24 & 58.64 & 52.63 & 32.54 & 34.28 & 25.08 & 26.44 & 32.00 & 91.56 & 40.85 & 10.20 & 100.00 & 51.62 & 44.36 & \textbf{45.87} \\
     & 0.50 & 30.39 & 46.56 & 55.62 & 58.99 & 52.54 & 33.01 & 32.77 & 24.85 & 25.52 & 35.50 & 91.88 & 40.24 & 12.15 & 100.00 & 50.73 & 43.54 & \textbf{45.89} \\
     & 0.75 & 29.79 & 44.40 & 54.79 & 58.79 & 50.49 & 32.63 & 29.65 & 24.28 & 23.37 & 40.00 & 91.78 & 40.54 & 10.63 & 100.00 & 52.01 & 45.05 & \textbf{45.51} \\
     & 0.90 & 31.04 & 38.72 & 52.98 & 56.19 & 45.95 & 27.55 & 26.02 & 23.57 & 19.73 & 36.00 & 91.09 & 39.70 & 9.55 & 99.50 & 48.65 & 47.78 & 43.38 \\
    \cmidrule(r){1-19}
     \multirow{4}{5em}{SnapKV\\ + VECTOR} & 0.25 & 31.31 & 47.23 & 55.62 & 59.42 & 51.19 & 31.27 & 34.64 & 25.01 & 26.69 & 27.00 & 91.88 & 40.13 & 10.65 & 100.00 & 50.97 & 44.30 & 45.46 \\
     & 0.50 & 31.93 & 47.52 & 56.78 & 59.48 & 51.04 & 31.02 & 33.11 & 25.09 & 26.19 & 25.00 & 91.95 & 38.49 & 10.60 & 99.50 & 48.56 & 44.94 & 45.08 \\
     & 0.75 & 31.96 & 46.80 & 56.70 & 58.39 & 50.12 & 30.47 & 31.14 & 24.78 & 24.13 & 33.00 & 91.71 & 39.67 & 9.63 & 99.50 & 48.44 & 45.32 & 45.11 \\
     & 0.90 & 30.36 & 43.94 & 54.65 & 55.96 & 48.54 & 31.06 & 28.17 & 24.91 & 22.17 & 42.50 & 90.15 & 40.00 & 11.50 & 99.50 & 50.69 & 49.06 & \textbf{45.20} \\
    \cmidrule(r){1-19}
     \multirow{4}{5em}{KVzip} & 0.25 & 30.88 & 48.12 & 56.35 & 59.01 & 49.82 & 32.53 & 34.58 & 25.27 & 26.81 & 19.50 & 91.37 & 38.04 & 11.38 & 100.00 & 54.46 & 45.59 & 45.23 \\
     & 0.50 & 30.87 & 47.64 & 57.01 & 55.67 & 49.32 & 28.62 & 35.14 & 25.14 & 27.27 & 15.50 & 88.53 & 34.42 & 10.19 & 100.00 & 50.76 & 44.97 & 43.82 \\
     & 0.75 & 32.38 & 45.70 & 54.79 & 56.35 & 47.38 & 25.80 & 34.35 & 25.16 & 26.64 & 46.00 & 86.48 & 32.59 & 11.80 & 99.00 & 49.41 & 45.42 & 44.95 \\
     & 0.90 & 29.93 & 40.28 & 53.83 & 52.65 & 39.23 & 26.38 & 31.77 & 23.68 & 25.66 & 45.00 & 90.63 & 37.69 & 10.50 & 59.50 & 42.06 & 50.83 & 41.23 \\
    \cmidrule(r){1-19}
     \multirow{4}{5em}{KVzip\\ + VECTOR} & 0.25 & 32.45 & 49.88 & 56.76 & 58.51 & 50.08 & 33.50 & 34.58 & 25.14 & 27.18 & 18.50 & 91.37 & 38.80 & 12.29 & 100.00 & 53.98 & 44.79 & \textbf{45.49} \\
     & 0.50 & 32.32 & 48.04 & 57.55 & 56.20 & 52.02 & 34.30 & 33.73 & 24.85 & 26.71 & 15.50 & 92.64 & 39.55 & 10.04 & 99.00 & 53.48 & 45.36 & \textbf{45.08} \\
     & 0.75 & 32.16 & 47.06 & 55.82 & 56.23 & 50.47 & 29.46 & 33.55 & 24.86 & 26.78 & 39.00 & 87.76 & 35.06 & 10.25 & 99.50 & 49.11 & 44.70 & \textbf{45.11} \\
     & 0.90 & 30.18 & 44.07 & 55.41 & 53.23 & 46.03 & 28.48 & 31.65 & 24.11 & 25.70 & 58.00 & 90.41 & 39.33 & 10.25 & 88.00 & 49.18 & 49.24 & \textbf{45.20} \\
    \cmidrule(r){1-19}
     \multirow{4}{5em}{PyramidKV} & 0.25 & 31.37 & 48.20 & 56.60 & 55.38 & 53.59 & 28.34 & 34.61 & 25.00 & 26.54 & 53.00 & 93.44 & 40.79 & 12.50 & 100.00 & 52.28 & 49.40 & 47.56 \\
     & 0.50 & 30.63 & 46.84 & 56.98 & 56.00 & 52.77 & 26.87 & 32.01 & 25.26 & 25.05 & 50.00 & 93.44 & 40.65 & 12.50 & 100.00 & 53.11 & 49.48 & 46.97 \\
     & 0.75 & 30.80 & 44.30 & 55.63 & 55.77 & 52.30 & 28.99 & 29.21 & 24.64 & 23.37 & 44.00 & 93.14 & 40.18 & 12.00 & 100.00 & 52.84 & 48.79 & 46.00 \\
     & 0.90 & 30.38 & 38.20 & 52.49 & 55.83 & 45.69 & 27.28 & 25.54 & 23.73 & 19.77 & 36.00 & 92.77 & 39.98 & 11.55 & 99.50 & 48.78 & 48.85 & 43.52 \\
    \cmidrule(r){1-19}
     \multirow{4}{5em}{PyramidKV\\ + VECTOR} & 0.25 & 30.67 & 48.51 & 57.72 & 55.66 & 54.38 & 28.68 & 34.74 & 25.36 & 26.45 & 51.50 & 93.37 & 40.34 & 12.50 & 99.50 & 53.29 & 49.36 & \textbf{47.63} \\
     & 0.50 & 31.87 & 49.23 & 57.69 & 55.47 & 52.18 & 27.39 & 31.71 & 25.03 & 24.83 & 54.50 & 93.37 & 39.07 & 12.00 & 99.50 & 53.52 & 49.61 & \textbf{47.31} \\
     & 0.75 & 30.23 & 45.69 & 56.14 & 57.11 & 50.11 & 26.61 & 29.50 & 24.81 & 22.59 & 56.00 & 93.37 & 40.22 & 11.00 & 99.50 & 52.78 & 48.99 & \textbf{46.54} \\
     & 0.90 & 28.85 & 43.13 & 54.43 & 56.22 & 47.83 & 27.24 & 26.96 & 24.29 & 20.47 & 48.50 & 93.45 & 40.69 & 10.00 & 99.50 & 50.85 & 48.51 & \textbf{45.06} \\
     
    \midrule
    
    \textbf{Qwen3-14B} & No Comp. & 30.39 & 43.89 & 52.76 & 62.30 & 56.04 & 33.33 & 32.74 & 24.43 & 24.98 & 69.50 & 90.93 & 41.87 & 8.75 & 99.42 & 66.16 & 63.87 & 50.08 \\
    \cmidrule(r){1-19}
     \multirow{4}{5em}{KeyDiff} & 0.25 & 27.78 & 36.58 & 49.92 & 51.68 & 45.59 & 27.28 & 32.95 & 24.19 & 24.74 & 69.50 & 87.77 & 41.12 & 13.50 & 98.08 & 59.37 & 55.65 & 46.61 \\
     & 0.50 & 25.05 & 30.04 & 45.26 & 40.15 & 35.99 & 21.38 & 28.82 & 23.19 & 22.84 & 56.50 & 85.77 & 38.79 & 12.00 & 94.83 & 40.39 & 50.48 & 40.72 \\
     & 0.75 & 21.53 & 21.56 & 34.26 & 25.28 & 28.31 & 14.90 & 20.24 & 21.20 & 15.69 & 31.50 & 84.27 & 34.69 & 10.00 & 83.00 & 20.63 & 48.67 & 32.23 \\
     & 0.90 & 12.52 & 11.94 & 24.86 & 19.12 & 24.55 & 5.80 & 11.83 & 19.78 & 6.88 & 2.50 & 83.50 & 30.54 & 10.00 & 39.00 & 11.26 & 49.31 & 22.71 \\
    \cmidrule(r){1-19}
     \multirow{4}{5em}{KeyDiff\\ + VECTOR} & 0.25 & 29.85 & 41.81 & 52.99 & 60.60 & 52.18 & 34.47 & 31.79 & 24.05 & 24.70 & 71.50 & 90.93 & 42.05 & 7.50 & 99.75 & 64.24 & 63.29 & \textbf{49.48} \\
     & 0.50 & 29.44 & 40.32 & 50.18 & 58.68 & 46.87 & 34.74 & 29.92 & 23.71 & 23.47 & 68.50 & 91.12 & 39.06 & 6.62 & 99.75 & 61.95 & 59.66 & \textbf{47.75} \\
     & 0.75 & 25.18 & 29.74 & 37.92 & 44.52 & 36.72 & 23.61 & 25.13 & 22.39 & 18.95 & 55.50 & 90.66 & 37.42 & 10.55 & 98.53 & 49.01 & 56.28 & \textbf{41.38} \\
     & 0.90 & 21.13 & 17.32 & 27.50 & 33.18 & 27.48 & 12.47 & 20.27 & 19.17 & 13.59 & 15.92 & 90.07 & 35.49 & 8.06 & 83.71 & 37.79 & 55.94 & \textbf{32.44} \\
    \cmidrule(r){1-19}
     \multirow{4}{5em}{SnapKV} & 0.25 & 30.30 & 43.48 & 52.83 & 62.09 & 55.71 & 33.29 & 32.39 & 23.91 & 24.72 & 71.00 & 90.93 & 42.02 & 8.25 & 99.42 & 66.37 & 64.28 & \textbf{50.06} \\
     & 0.50 & 30.37 & 43.59 & 52.53 & 61.95 & 54.06 & 34.46 & 31.23 & 24.10 & 23.61 & 70.50 & 91.10 & 41.90 & 8.50 & 98.83 & 66.34 & 63.81 & \textbf{49.81} \\
     & 0.75 & 31.13 & 41.17 & 51.30 & 62.81 & 53.77 & 36.23 & 29.65 & 24.25 & 21.40 & 67.75 & 91.10 & 41.46 & 7.20 & 98.58 & 64.52 & 63.73 & \textbf{49.13} \\
     & 0.90 & 30.47 & 36.76 & 50.08 & 63.57 & 48.37 & 36.04 & 24.56 & 23.25 & 17.75 & 48.00 & 91.44 & 40.83 & 9.62 & 97.50 & 60.91 & 61.56 & 46.29 \\
    \cmidrule(r){1-19}
     \multirow{4}{5em}{SnapKV\\ + VECTOR} & 0.25 & 28.73 & 43.50 & 53.01 & 61.57 & 55.37 & 32.46 & 32.32 & 24.13 & 24.78 & 70.50 & 91.06 & 41.81 & 7.35 & 99.42 & 66.39 & 64.27 & 49.79 \\
     & 0.50 & 28.89 & 43.00 & 51.70 & 62.67 & 53.38 & 35.57 & 31.33 & 23.91 & 23.91 & 68.50 & 91.02 & 38.60 & 7.22 & 98.47 & 65.79 & 60.16 & 49.01 \\
     & 0.75 & 28.64 & 43.39 & 50.61 & 62.07 & 53.29 & 35.73 & 30.73 & 23.60 & 23.05 & 65.50 & 91.02 & 38.92 & 7.81 & 97.75 & 65.35 & 60.63 & 48.63 \\
     & 0.90 & 29.86 & 42.65 & 50.21 & 61.96 & 52.55 & 36.31 & 28.16 & 23.50 & 20.17 & 57.42 & 90.41 & 40.08 & 7.10 & 96.29 & 63.62 & 62.13 & \textbf{47.65} \\
    \cmidrule(r){1-19}
     \multirow{4}{5em}{KVzip} & 0.25 & 30.23 & 43.49 & 52.16 & 62.23 & 54.77 & 33.32 & 32.69 & 24.48 & 24.83 & 70.00 & 91.10 & 41.62 & 7.95 & 99.17 & 66.22 & 64.12 & \textbf{49.90} \\
     & 0.50 & 30.59 & 43.36 & 51.93 & 62.58 & 54.61 & 33.39 & 32.91 & 24.71 & 24.87 & 71.50 & 91.10 & 41.85 & 9.55 & 97.17 & 65.16 & 64.35 & \textbf{49.98} \\
     & 0.75 & 31.01 & 43.70 & 52.81 & 63.94 & 53.10 & 33.32 & 32.50 & 24.99 & 24.85 & 73.50 & 91.08 & 40.41 & 8.53 & 92.95 & 61.53 & 63.58 & \textbf{49.49} \\
     & 0.90 & 28.14 & 38.72 & 52.82 & 62.11 & 44.31 & 29.59 & 30.44 & 24.26 & 23.96 & 62.00 & 89.83 & 37.87 & 6.97 & 88.33 & 34.63 & 61.98 & 44.75 \\
    \cmidrule(r){1-19}
     \multirow{4}{5em}{KVzip\\ + VECTOR} & 0.25 & 29.77 & 43.44 & 52.45 & 60.91 & 54.52 & 31.30 & 32.58 & 24.31 & 25.11 & 69.00 & 91.06 & 41.60 & 7.85 & 100.00 & 65.94 & 63.49 & 49.58 \\
     & 0.50 & 29.26 & 42.33 & 51.80 & 61.51 & 51.89 & 34.80 & 31.51 & 23.94 & 24.49 & 67.00 & 90.87 & 40.47 & 7.97 & 99.75 & 63.35 & 60.16 & 48.82 \\
     & 0.75 & 29.93 & 43.24 & 51.75 & 62.07 & 51.65 & 32.54 & 31.49 & 24.04 & 24.36 & 72.50 & 90.62 & 40.06 & 9.70 & 97.53 & 63.24 & 61.45 & 49.14 \\
     & 0.90 & 31.85 & 39.00 & 53.82 & 60.74 & 47.52 & 33.74 & 30.05 & 24.14 & 24.01 & 67.00 & 91.47 & 36.60 & 7.84 & 97.67 & 53.09 & 61.87 & \textbf{47.53} \\
    \cmidrule(r){1-19}
     \multirow{4}{5em}{PyramidKV} & 0.25 & 30.22 & 44.72 & 53.48 & 64.23 & 57.14 & 37.73 & 32.46 & 24.01 & 24.99 & 73.00 & 90.43 & 42.39 & 9.56 & 98.27 & 66.81 & 63.77 & \textbf{50.83} \\
     & 0.50 & 29.71 & 41.64 & 51.70 & 64.46 & 55.08 & 38.02 & 30.50 & 23.78 & 22.10 & 71.00 & 90.43 & 42.10 & 9.03 & 98.27 & 65.50 & 63.82 & \textbf{49.82} \\
     & 0.75 & 30.02 & 41.28 & 51.89 & 63.97 & 52.06 & 37.32 & 28.72 & 23.27 & 21.30 & 62.50 & 90.46 & 41.67 & 7.76 & 98.52 & 64.40 & 62.51 & 48.60 \\
     & 0.90 & 29.20 & 36.79 & 49.56 & 62.71 & 47.65 & 35.42 & 24.46 & 22.56 & 17.64 & 48.00 & 91.39 & 40.48 & 8.22 & 97.42 & 60.98 & 61.72 & \textbf{45.89} \\
    \cmidrule(r){1-19}
     \multirow{4}{5em}{PyramidKV\\ + VECTOR} & 0.25 & 28.60 & 45.68 & 53.10 & 64.06 & 58.00 & 38.33 & 32.59 & 23.45 & 24.61 & 74.00 & 90.71 & 40.64 & 8.87 & 98.04 & 66.26 & 64.04 & 50.69 \\
     & 0.50 & 28.88 & 44.03 & 51.02 & 63.38 & 53.83 & 37.04 & 30.25 & 23.27 & 21.90 & 70.50 & 90.97 & 39.10 & 8.40 & 98.00 & 65.34 & 62.77 & 49.29 \\
     & 0.75 & 29.80 & 42.22 & 50.31 & 64.47 & 53.41 & 36.16 & 28.86 & 23.41 & 20.30 & 67.50 & 90.97 & 39.71 & 8.93 & 97.00 & 63.44 & 62.75 & \textbf{48.70} \\
     & 0.90 & 28.06 & 38.43 & 48.11 & 63.53 & 50.95 & 33.56 & 26.06 & 23.25 & 17.12 & 47.00 & 91.02 & 39.06 & 7.46 & 95.08 & 59.21 & 61.13 & 45.56 \\

    \bottomrule
  \end{tabular}
  }
  \caption{LongBench results for four eviction baselines and their VECTOR-augmented variants under compression ratios $p_c \in \{0.25,0.50,0.75,0.90\}$. Approximation ratios are set by Eq.~\ref{eq:p_a_deploy}. \rev{Table~\ref{tab:exp_main} summarizes these results. Results for Qwen3-0.6B are in Table~\ref{tab:exp_qwen06b}.}}
  \label{tab:exp_main_full}
\end{table*}

\begin{table*}[t]
\centering
\resizebox{\textwidth}{!}{
\begin{tabular}{lcccccccccccccccccc}
    \toprule
    & & \multicolumn{16}{c}{LongBench} & \\
    \cmidrule(r){3-18}
    Method & \makecell{Comp. \\Ratio} & narrativeqa & qasper & \makecell{multifield- \\ qa\_en} & hotpotqa & 2wikimqa & musique & gov\_report & qmsum & multi\_news & trec & triviaqa & samsum & \makecell{passage\_ \\ count} & \makecell{passage\_ret- \\ rieval\_en} & lcc & repobench-p & Avg. \\
    
    \midrule

    \textbf{Qwen3-0.6B} & No Comp. & 15.71 & 18.59 & 46.77 & 35.60 & 29.52 & 14.47 & 27.06 & 20.52 & 24.36 & 20.00 & 70.42 & 35.13 & 0.50 & 75.00 & 31.79 & 29.42 & 30.93 \\
    \cmidrule(r){1-19}
     \multirow{4}{5em}{KeyDiff} & 0.25 & 15.20 & 15.83 & 36.94 & 25.09 & 26.05 & 7.78 & 25.86 & 19.74 & 23.02 & 54.50 & 66.70 & 33.17 & 0.62 & 37.50 & 29.63 & 31.36 & 28.06 \\
     & 0.50 & 12.09 & 11.47 & 29.91 & 16.70 & 24.98 & 5.66 & 20.06 & 19.24 & 17.83 & 46.00 & 62.76 & 32.18 & 0.28 & 15.00 & 25.83 & 30.16 & 23.13 \\
     & 0.75 & 10.30 & 7.71 & 25.17 & 12.66 & 19.48 & 4.29 & 11.89 & 18.19 & 10.30 & 12.50 & 59.62 & 28.04 & 0.50 & 6.00 & 13.41 & 27.41 & 16.72 \\
     & 0.90 & 6.27 & 4.83 & 21.81 & 9.80 & 20.29 & 3.69 & 6.36 & 18.12 & 5.97 & 1.50 & 52.63 & 22.28 & 0.85 & 5.00 & 7.64 & 26.70 & 13.36 \\
    \cmidrule(r){1-19}
     \multirow{4}{5em}{KeyDiff\\ + VECTOR} & 0.25 & 17.07 & 18.56 & 45.54 & 32.10 & 31.04 & 11.36 & 25.73 & 20.54 & 23.14 & 27.00 & 68.89 & 33.75 & 1.00 & 68.00 & 30.52 & 29.69 & \textbf{30.25} \\
     & 0.50 & 18.00 & 16.59 & 38.60 & 29.27 & 27.86 & 11.42 & 23.28 & 20.44 & 21.59 & 38.50 & 69.47 & 34.18 & 1.50 & 54.00 & 27.67 & 30.21 & \textbf{28.91} \\
     & 0.75 & 14.85 & 12.53 & 26.85 & 20.08 & 21.58 & 7.00 & 16.91 & 19.61 & 15.21 & 33.50 & 67.99 & 34.05 & 1.00 & 19.00 & 25.01 & 30.86 & \textbf{22.88} \\
     & 0.90 & 8.80 & 13.43 & 23.16 & 16.86 & 26.93 & 5.22 & 12.82 & 18.96 & 10.75 & 14.00 & 64.78 & 32.94 & 1.53 & 5.00 & 16.55 & 31.33 & \textbf{18.94} \\
    \cmidrule(r){1-19}
     \multirow{4}{5em}{SnapKV} & 0.25 & 15.41 & 18.50 & 46.66 & 37.12 & 30.04 & 14.66 & 26.88 & 20.56 & 23.39 & 20.00 & 70.61 & 33.82 & 1.00 & 74.50 & 30.58 & 29.06 & 30.80 \\
     & 0.50 & 15.69 & 18.31 & 46.54 & 36.82 & 30.01 & 13.99 & 25.06 & 20.63 & 21.99 & 30.00 & 70.32 & 33.49 & 0.50 & 75.00 & 31.12 & 29.37 & \textbf{31.18} \\
     & 0.75 & 15.98 & 17.09 & 45.31 & 37.13 & 30.28 & 12.81 & 22.15 & 20.17 & 19.05 & 30.50 & 70.68 & 32.56 & 0.50 & 77.00 & 34.00 & 31.66 & \textbf{31.05} \\
     & 0.90 & 16.82 & 15.85 & 39.96 & 37.25 & 29.73 & 12.74 & 17.53 & 19.39 & 14.46 & 23.50 & 71.89 & 32.24 & 1.00 & 75.50 & 35.61 & 34.17 & 29.85 \\
    \cmidrule(r){1-19}
     \multirow{4}{5em}{SnapKV\\ + VECTOR} & 0.25 & 15.76 & 18.80 & 46.88 & 36.99 & 29.56 & 14.54 & 26.98 & 20.55 & 23.85 & 21.50 & 70.76 & 34.53 & 0.50 & 74.00 & 31.05 & 29.57 & \textbf{30.99} \\
     & 0.50 & 15.78 & 17.98 & 46.92 & 35.64 & 28.64 & 15.20 & 26.69 & 20.68 & 23.65 & 19.00 & 70.40 & 34.75 & 0.50 & 75.00 & 28.32 & 28.42 & 30.47 \\
     & 0.75 & 16.79 & 17.14 & 45.31 & 35.38 & 30.64 & 13.90 & 25.39 & 20.58 & 22.12 & 29.50 & 69.87 & 35.65 & 0.50 & 76.00 & 27.99 & 28.62 & 30.96 \\
     & 0.90 & 17.57 & 16.33 & 45.26 & 36.83 & 27.15 & 13.98 & 23.39 & 20.09 & 19.34 & 37.00 & 69.70 & 34.63 & 0.50 & 78.00 & 32.79 & 31.15 & \textbf{31.48} \\
    \cmidrule(r){1-19}
     \multirow{4}{5em}{KVzip} & 0.25 & 15.95 & 18.26 & 46.78 & 36.45 & 29.65 & 14.61 & 27.21 & 20.32 & 24.14 & 20.00 & 70.10 & 34.67 & 1.00 & 74.00 & 30.88 & 29.22 & 30.83 \\
     & 0.50 & 15.63 & 16.92 & 48.64 & 35.34 & 30.56 & 14.86 & 27.27 & 20.23 & 24.28 & 27.00 & 70.68 & 34.53 & 0.50 & 75.00 & 30.64 & 30.12 & \textbf{31.39} \\
     & 0.75 & 13.77 & 14.41 & 45.42 & 33.25 & 31.32 & 10.66 & 26.18 & 20.37 & 24.46 & 54.50 & 69.72 & 33.93 & 0.50 & 28.50 & 31.48 & 30.39 & 29.30 \\
     & 0.90 & 12.74 & 15.93 & 37.63 & 18.74 & 28.31 & 5.90 & 23.55 & 19.79 & 22.06 & 45.50 & 67.34 & 30.21 & 0.00 & 11.00 & 20.82 & 28.95 & 24.28 \\
    \cmidrule(r){1-19}
     \multirow{4}{5em}{KVzip\\ + VECTOR} & 0.25 & 16.03 & 18.99 & 46.46 & 35.73 & 28.43 & 14.15 & 27.33 & 20.67 & 23.92 & 23.00 & 70.17 & 35.36 & 0.50 & 75.00 & 30.23 & 29.29 & \textbf{30.95} \\
     & 0.50 & 15.68 & 17.71 & 45.41 & 33.80 & 29.33 & 12.52 & 26.50 & 20.86 & 23.77 & 21.50 & 70.03 & 35.96 & 0.50 & 70.00 & 27.05 & 28.82 & 29.96 \\
     & 0.75 & 15.66 & 16.15 & 47.52 & 34.00 & 30.30 & 13.68 & 26.56 & 20.73 & 24.19 & 44.00 & 69.64 & 35.11 & 0.50 & 69.00 & 26.82 & 29.32 & \textbf{31.45} \\
     & 0.90 & 14.08 & 15.68 & 42.87 & 26.00 & 30.16 & 6.46 & 25.40 & 20.42 & 23.35 & 55.00 & 69.76 & 33.23 & 0.00 & 15.50 & 25.10 & 29.76 & \textbf{27.05} \\
    \cmidrule(r){1-19}
     \multirow{4}{5em}{PyramidKV} & 0.25 & 15.69 & 15.24 & 46.01 & 29.62 & 24.43 & 11.38 & 25.90 & 21.33 & 22.96 & 4.50 & 62.82 & 16.65 & 3.00 & 67.00 & 27.47 & 25.87 & 26.24 \\
     & 0.50 & 16.12 & 13.47 & 44.18 & 30.51 & 23.60 & 8.34 & 22.04 & 20.82 & 18.66 & 4.00 & 62.86 & 20.80 & 3.00 & 65.00 & 26.74 & 27.22 & 25.46 \\
     & 0.75 & 16.70 & 15.09 & 42.33 & 30.35 & 24.68 & 9.49 & 19.84 & 19.87 & 18.81 & 17.50 & 66.89 & 24.97 & 3.00 & 63.50 & 33.45 & 27.77 & \textbf{27.14} \\
     & 0.90 & 16.75 & 16.55 & 38.99 & 30.78 & 30.00 & 7.29 & 16.98 & 19.34 & 14.46 & 23.50 & 70.72 & 30.03 & 1.00 & 68.50 & 35.55 & 32.26 & \textbf{28.29} \\
    \cmidrule(r){1-19}
     \multirow{4}{5em}{PyramidKV\\ + VECTOR} & 0.25 & 16.60 & 15.89 & 45.92 & 28.82 & 24.79 & 10.63 & 26.12 & 20.65 & 23.24 & 4.50 & 63.25 & 18.51 & 3.00 & 67.00 & 26.20 & 25.88 & \textbf{26.31} \\
     & 0.50 & 17.17 & 13.72 & 44.70 & 29.50 & 24.08 & 9.85 & 24.32 & 21.07 & 20.82 & 10.50 & 63.93 & 19.45 & 2.50 & 67.50 & 25.39 & 25.82 & \textbf{26.27} \\
     & 0.75 & 16.61 & 13.30 & 42.67 & 29.29 & 23.38 & 7.99 & 22.27 & 20.32 & 17.76 & 12.50 & 61.41 & 18.82 & 3.00 & 65.50 & 23.90 & 25.88 & 25.29 \\
     & 0.90 & 16.15 & 14.89 & 37.49 & 27.55 & 19.77 & 7.92 & 19.17 & 19.80 & 13.94 & 13.50 & 60.80 & 21.43 & 3.00 & 63.50 & 23.73 & 26.10 & 24.30 \\
    \bottomrule
\end{tabular}
}
\caption{LongBench results for four eviction baselines and their 
VECTOR-augmented variants on Qwen3-0.6B, under compression 
ratios $p_c \in \{0.25, 0.50, 0.75, 0.90\}$. Approximation 
ratios are set by Eq.~\ref{eq:p_a_deploy}. Experimental setup 
follows Table~\ref{tab:exp_main_full}.}
\label{tab:exp_qwen06b}
\end{table*}

\section{Theoretical Analysis}\label{sec:theory}

% Since the proposed pipeline introduces the additional \textit{Approximation} state,
% we analyze how the approximation performance and the importance score distribution
% jointly affect the overall distortion\pf{How to understand this 'distortion'} after compression
% (Proposition~\ref{prop:skewness_threshold}), and provide an illustrative example
% of $R^2_\textrm{approx}$ \pf{Why R2? Any high-level interpretation? I think the connection of this section is not strong enough. Maybe we mention some theoretical findings at the beginning?} under a Gaussian residual assumption
% (Example~\ref{expl:gaussian_distribution}).

% To quantify the distortion, we define $\mathcal{E}$ as the importance-weighted
% proportion of signal loss \pf{this is the first time this term comes out, any explanation?}. Specifically, an evicted token contributes a full signal
% loss of $1$, a retained token contributes $0$, and an approximated token
% contributes its normalized reconstruction error
% ${\|V_i-\hat{V}_i\|_2^2}/{\|V_i\|_2^2} \in [0,1]$.

We analyze the effect of introducing the \textit{Approximation} state under a fixed memory budget. The central question is: under what conditions does expanding the \textit{Approximation} tier reduce information loss, compared to binary eviction?

To formalize this, we define $\mathcal{E}$ as an importance-weighted measure of signal loss after compression. \rev{Each token contributes its signal loss weighted by its importance score $w_i$. The signal loss is $1$ for an evicted token, $0$ for a retained token, and the normalized reconstruction error ${\|V_i-\hat{V}_i\|_2^2}/{\|V_i\|_2^2} \in [0,1]$ for an approximated token.} A smaller $\mathcal{E}$ means less distortion to the attention computation.

We then characterize when expanding the \textit{Approximation} state reduces $\mathcal{E}$ (Proposition~\ref{prop:skewness_threshold}). The key quantity is $R^2_\textrm{approx}$, the OLS prediction quality measured specifically over the tokens selected for approximation. Intuitively, $R^2_\textrm{approx}$ close to $1$ means approximated tokens are well-reconstructed, contributing little to $\mathcal{E}$. We derive a threshold on $R^2_\textrm{approx}$ that depends on the skewness of the importance score distribution. We also provide a closed-form expression for $\mathcal{E}$ under a Gaussian residual assumption (Example~\ref{expl:gaussian_distribution}).

Let $\mathcal{I}_E(p_a,p_c)$ and $\mathcal{I}_A(p_a,p_c)$ denote the index sets of evicted and approximated tokens, respectively, and let $w_i$ denote the importance score of the $i$-th token. We then write $\mathcal{E}$ as:
\begin{align*}
    \mathcal{E}(p_a,p_c)
    &= \sum_{i\in \mathcal{I}_E(p_a,p_c)} w_i \\
    &\quad + \sum_{i\in \mathcal{I}_A(p_a,p_c)} w_i
          \frac{\|V_i-\hat{V}_i\|_2^2}{\|V_i\|^2_2} \\
    &\approx \sum_{i\in \mathcal{I}_E(p_a,p_c)} w_i
        + \Bigl(\sum_{i\in \mathcal{I}_A(p_a,p_c)} w_i\Bigr)\, \mu_A,
\end{align*}
where the approximation assumes that the per-token reconstruction errors
${\|V_i-\hat{V}_i\|_2^2}/{\|V_i\|_2^2}$ are approximately independent of the
importance scores $w_i$, and $\mu_A = 1 - R^2_\textrm{approx}(p_a,p_c)$ denotes
the mean relative reconstruction error over $\mathcal{I}_A$. Here,
$R^2_\textrm{approx}(p_a,p_c)$ is the $R^2$ computed solely over the tokens in
the \textit{Approximation} set.

% [REV-DEL] The following proposition characterizes the relationship among
% [REV-DEL] $R^2_\textrm{approx}$, the importance score distribution, and the overall
% [REV-DEL] distortion $\mathcal{E}$. Specifically, we derive the condition on
% [REV-DEL] $R^2_\textrm{approx}$ under which expanding the \textit{Approximation} state
% [REV-DEL] reduces $\mathcal{E}$, and show that this condition depends on the skewness of
% [REV-DEL] the importance score distribution.
% [REV-DEL] 
% [REV-DEL] \begin{proposition}\label{prop:skewness_threshold}
% [REV-DEL] Let $\bar{w} = \left(1-\sum_{i\in\mathcal{I}_E}w_i\right)/(1-p_c+p_a)$ be the
% [REV-DEL] average importance score of the tokens in the expanded pool, and let $w^*$ be
% [REV-DEL] the importance score of the token at the truncation boundary of the
% [REV-DEL] \textit{Eviction} set. Expanding the \textit{Approximation} state reduces the
% [REV-DEL] global distortion if:
% [REV-DEL] \begin{equation*}
% [REV-DEL]     R^2_\textrm{approx} > \frac{\bar{w}}{w^*+\bar{w}}.
% [REV-DEL] \end{equation*}
% [REV-DEL] \end{proposition}
\rev{Under this approximation, Proposition~\ref{prop:skewness_threshold} (Section~\ref{sec:method_pipeline}) gives the condition on $R^2_\textrm{approx}$ under which expanding the \textit{Approximation} state reduces $\mathcal{E}$. There, $\bar{w} = \left(1-\sum_{i\in\mathcal{I}_E}w_i\right)/(1-p_c+p_a)$ is the average importance score of the tokens in the expanded pool, and $w^*$ is the importance score of the token at the truncation boundary of the \textit{Eviction} set. The proof is given in Appendix~\ref{sec:appendix:proof:prop}.}

Based on Proposition~\ref{prop:skewness_threshold}, the required predictability
$R^2_\textrm{approx}$ is governed by the skewness of the importance score
distribution. 
% \yue{When increasing $p_a$, tokens are moved from \textit{Eviction} set and the \textit{Retention} set to the \textit{Approximation} set. While the boundary tokens moved from the \textit{Eviction} set to the \textit{Approximation}/\textit{Retention} set can be better recovered, the other boundary tokens moved from the \textit{Retention} set to the \textit{Approximation} set suffers an increase in distortion.} Therefore, for highly skewed distributions where $\bar{w} \gg w^*$ (i.e.,
% the importance scores of evicted tokens are negligibly small), 
% the OLS approximation must achieve higher accuracy to \yue{compensate for the latter tokens and} yield a net reduction in distortion.
% When increasing $p_a$, tokens are moved from \textit{Eviction} set and the \textit{Retention} set to the \textit{Approximation} set. While the boundary tokens moved from the \textit{Eviction} set to the \textit{Approximation}/\textit{Retention} set can be better recovered, the other boundary tokens moved from the \textit{Retention} set to the \textit{Approximation} set suffers an increase in distortion. Therefore, for highly skewed distributions where $\bar{w} \gg w^*$ (i.e., the importance scores of evicted tokens are negligibly small), the OLS approximation must achieve higher accuracy to compensate for the latter tokens and yield a net reduction in distortion.
When expanding the \textit{Approximation} tier, tokens enter from both the \textit{Eviction} set (gaining recovery) and the \textit{Retention} set (incurring approximation error). For highly skewed distributions where $\bar{w} \gg w^*$, the latter effect dominates, and OLS must achieve higher accuracy to compensate and yield a net reduction in distortion.

Besides the above, the following example adopts a Gaussian assumption on the OLS
residuals to examine how the shape of the residual distribution affects
$R^2_\textrm{approx}$ and, consequently, $\mathcal{E}$:

\begin{example}\label{expl:gaussian_distribution}
    Let $\phi$ and $\Phi$ denote the PDF and CDF of the standard normal
    distribution, respectively. Let $r_i$ denote the signed OLS residual for the
    $i$-th token, satisfying $r_i^2 = \|V_i - \hat{V}_i\|_2^2 = \varepsilon_i$.
    Assume $r_i \overset{\mathrm{i.i.d.}}{\sim} \mathcal{N}(0,\sigma^2)$, and
    let $\Sigma^2$ denote the variance of $\|V_i\|_2$ ($\Sigma^2>\sigma^2$).
    Define
    $\eta(p_a,p_c) := \sigma\,\Phi^{-1}\!\left(1/2 + p_a/(1-p_c+p_a)\right)$.
    Then, given a target retention ratio $1-p_c$:
    \begin{align*}
        \mathcal{E}(p_a,p_c) &\approx
            \sum_{i\in\mathcal{I}_E(p_a,p_c)} w_i \\
            &\quad + 2p_a\bar{w}\,\bigl(1-R^2_\textrm{approx}(p_a,p_c)\bigr),
    \end{align*}
    with
    \[
        1-R^2_\textrm{approx} = \frac{\sigma^2}{\Sigma^2}
        \Bigl(1 - \tfrac{1-p_c+p_a}{p_a}\,
        \tfrac{\eta}{\sigma}\,\phi\bigl(\tfrac{\eta}{\sigma}\bigr)\Bigr),
    \]
    where $\eta$ is short for $\eta(p_a,p_c)$. This shows that $1-R^2_\textrm{approx}(p_a,p_c)$ depends on $\sigma^2$ and
    $\Sigma^2$ under the Gaussian distribution.
\end{example}

Combining Proposition~\ref{prop:skewness_threshold} and
Example~\ref{expl:gaussian_distribution}, when the importance score distribution
concentrates on the \textit{Approximation} and \textit{Retention} sets (i.e.,
$\bar{w} \gg w^*$), the method is effective provided that the values are highly
predictable from the keys, i.e., $\sigma^2 \ll \Sigma^2$.

% For deployment, since the KV cache varies across samples, $w^*$ and $\bar{w}$
% change accordingly, making it impractical to determine $p_a$ dynamically. We
% therefore adopt the following empirical formula:
% \begin{equation} \label{eq:p_a_deploy}
%     p_a^{\mathrm{deploy}} = \min\!\left(p_c/2,\; (1-p_c-\epsilon)/2,\; 0.2\right),
% \end{equation}
% % where $\epsilon>0$ is a small constant ensuring a non-empty \textit{Retention}
% % tier, and the upper bound $0.2$ reflects the typical magnitude of the gap between
% % $w^*$ and $\bar{w}$ while keeping $p_a$ small.

% where $\epsilon>0$ is a small constant ensuring a non-empty \textit{Retention}
% tier, and the upper bound $0.2$ is an empirically determined conservative 
% threshold; we find that larger values of $p_a$ do not yield further 
% gains and may reduce performance as the \textit{Approximation} set 
% grows to include tokens with higher reconstruction errors.
% \yp{Doing this experiment again with 0.2 limitation removed. (only $p_c = 0.5$
% affected) If finished before ddl and result good, will remove 0.2 here}

% [REV] The deployment rule for p_a (Eq. eq:p_a_deploy) moved to Section 3.2 (method.tex).

\section{Proofs}\label{sec:appendix:proof}
\subsection{Proof of Proposition \ref{prop:skewness_threshold}}
\label{sec:appendix:proof:prop}

Given the formula for $\mathcal{E}$ and $\mu_A$, we have
\begin{multline}
    \mathcal{E}(p_a,p_c) \approx
        \sum_{i\in\mathcal{I}_E(p_a,p_c)} w_i \\
        + \Bigl(\sum_{i\in\mathcal{I}_A(p_a,p_c)} w_i\Bigr)
          \bigl(1 - R^2_\textrm{approx}(p_a,p_c)\bigr).
\end{multline}

Since $w^*$ is the importance score at the truncation boundary, $\mathcal{I}_E$
has proportion $p_c - p_a$, and $\mathcal{I}_A$ contains $2p_a$ proportion of
tokens, we have
\begin{equation}
    \sum_{i\in\mathcal{I}_E(p_a+\delta,p_c)} \!\! w_i
    - \sum_{i\in\mathcal{I}_E(p_a,p_c)} \!\! w_i
    \approx -w^*\delta,
\end{equation}
i.e., increasing $p_a$ by $\delta$ removes $\delta$ boundary tokens with weight
$w^*$ from $\mathcal{I}_E$. Similarly,
\begin{equation}
    \sum_{i\in\mathcal{I}_A(p_a+\delta,p_c)} \!\! w_i
    - \sum_{i\in\mathcal{I}_A(p_a,p_c)} \!\! w_i
    \approx (w^*+\bar{w})\delta,
\end{equation}
since $|\mathcal{I}_A|$ grows by $2\delta$: $\delta$ tokens enter from
$\mathcal{I}_E$ with weight $w^*$, and $\delta$ tokens enter from the
\textit{Retention} set with weight $\approx\bar{w}$. The latter holds because
the partition between \textit{Approximation} and \textit{Retention} is determined
by reconstruction error rather than importance score, so the transferred tokens
carry average weight $\bar{w}$.

Combining the above, we obtain:
\begin{multline}
    \mathcal{E}(p_a+\delta,p_c) - \mathcal{E}(p_a,p_c)
    \approx -w^*\delta \\
             + (w^*+\bar{w})\delta\bigl(1-R^2_\textrm{approx}(p_a,p_c)\bigr) \\
             - 2p_a\bar{w}\delta
               \frac{\partial R^2_\textrm{approx}}{\partial p_a}.
\end{multline}
For $\mathcal{E}(p_a+\delta,p_c) < \mathcal{E}(p_a,p_c)$ (with $\delta > 0$),
we require (writing $D := 2p_a\bar{w}\,\partial R^2_\textrm{approx}/\partial p_a$ for brevity):
\begin{align}
    &-w^* + (w^*+\bar{w})\bigl(1-R^2_\textrm{approx}\bigr) - D < 0 \\
    \Leftrightarrow\;
       &(w^*+\bar{w})\bigl(1-R^2_\textrm{approx}\bigr) < w^* + D \\
    \Leftrightarrow\;
       &1 - R^2_\textrm{approx} < \frac{w^* + D}{w^*+\bar{w}} \\
    \Leftrightarrow\;
       &R^2_\textrm{approx} > 1 - \frac{w^* + D}{w^*+\bar{w}}.
\end{align}
For small $p_a$, the term $D$
is negligible, yielding:
$$R^2_\textrm{approx}(p_a,p_c)
  > 1 - \frac{w^*}{w^*+\bar{w}}
  = \frac{\bar{w}}{w^*+\bar{w}}. \qed$$

\subsection{Proof of Example \ref{expl:gaussian_distribution}}
\label{sec:appendix:proof:example}

Since the selection of $\mathcal{I}_E(p_a,p_c)$ is independent of the
reconstruction error, the signed residuals $r_i$ of non-evicted tokens follow
the same $\mathcal{N}(0,\sigma^2)$ distribution. Further invoking the
independence between $r_i^2$ and $w_i$, the overall distortion is:
\begin{align}
    \mathcal{E}(p_a,p_c)
    &= \sum_{i\in\mathcal{I}_E(p_a,p_c)} w_i
        + \frac{1}{\Sigma^2}
          \sum_{i\in\mathcal{I}_A(p_a,p_c)} w_i r_i^2 \\
    &\approx \sum_{i\in\mathcal{I}_E(p_a,p_c)} w_i \nonumber\\
    &\quad + \frac{2p_a\bar{w}}{\Sigma^2}
                \cdot \frac{\sum_{i\in\mathcal{I}_A(p_a,p_c)} r_i^2}
                           {|\mathcal{I}_A(p_a,p_c)|}.
\end{align}
Since the \textit{Approximation} set selects the $2p_a$ tokens with the smallest
reconstruction errors (i.e., those with $|r_i| \leq \eta(p_a,p_c)$),
corresponding to the central $2p_a/(1-p_c+p_a)$ fraction of the Gaussian
distribution, the conditional distribution of $r_i$ given
$i \in \mathcal{I}_A(p_a,p_c)$ is a truncated normal on
$[-\eta(p_a,p_c),\, \eta(p_a,p_c)]$, where
$\eta(p_a,p_c) = \sigma\Phi^{-1}(1/2 + p_a/(1-p_c+p_a))$, and
$2\Phi(\eta/\sigma)-1 = 2p_a/(1-p_c+p_a)$.

Recalling that for $X \sim \mathcal{N}(0,\sigma^2)$:
$$\mathbb{E}(X^2 \mid |X| \leq x)
  = \sigma^2\!\left[1 - \frac{2(x/\sigma)\phi(x/\sigma)}{2\Phi(x/\sigma)-1}
    \right],$$
we obtain, with $\eta$ short for $\eta(p_a,p_c)$:
\begin{align*}
    \mathcal{E}(p_a,p_c) &\approx
        \sum_{i\in\mathcal{I}_E(p_a,p_c)} w_i \\
        &\quad + 2p_a\bar{w}\,\bigl(1-R^2_\textrm{approx}(p_a,p_c)\bigr),
\end{align*}
where
\[
    1-R^2_\textrm{approx} = \frac{\sigma^2}{\Sigma^2}
    \Bigl(1 - \tfrac{1-p_c+p_a}{p_a}\,
    \tfrac{\eta}{\sigma}\,\phi\bigl(\tfrac{\eta}{\sigma}\bigr)\Bigr). \qed
\]

\section{Reproducibility Details}
\label{app:exp_repro}
% [REV-DEL] We provide implementation details for reproducibility and checklist reporting.
\rev{We provide implementation details for reproducibility.}

\subsection{LongBench Setup Details}
\begin{itemize}
    \item \textbf{Code base and framework.} All LongBench and NIAH experiments in this section use the KVPress evaluation framework and official baseline implementations~\citep{nvidia2025kvpress}.
    \item \textbf{Repository version.} KVPress commit hash:\\ {\scriptsize \texttt{91773cb8b5713fe6ad54e3e7a97dd343fb40d37b}}.
    \item \textbf{Model identifiers.} The evaluated models are \path{meta-llama/Llama-3.1-8B-Instruct}, \path{Qwen/Qwen3-14B}, and \path{Qwen/Qwen3-0.6B}.
% [REV-DEL]     \item \textbf{Calibration protocol.} For each model, we perform one-time offline OLS calibration on C4 dataset~\citep{dodge2021documenting}: 10,000 training sequences (length 4,096) are used to fit layer-wise $K\!\rightarrow\!V$ linear maps, and a held-out set of 100 sequences (length 4,096) is used for validation (Section~\ref{sec:approximation}). The calibrated $W_\mathrm{OLS}$ weights are fixed for all downstream experiments.
    \item \rev{\textbf{Calibration protocol.} For each model, we perform one-time offline OLS calibration on the C4 dataset~\citep{dodge2021documenting}. The weights used in all downstream experiments are fit on 5,000 sequences (Llama-3.1-8B) or 10,000 sequences (Qwen3-14B and Qwen3-0.6B) of 2,048 tokens. The cross-model predictability study in Table~\ref{tab:method_r2} uses 10,000 training sequences and a held-out set of 100 sequences, all of 4,096 tokens (Section~\ref{sec:approximation}). The calibrated $W_\mathrm{OLS}$ weights are fixed for all downstream experiments.}
    \item \textbf{Benchmark split and coverage.} For LongBench, we evaluate the full test split of the 16-task English/code subset (i.e., 21 original tasks minus 5 Chinese tasks), and report the arithmetic mean over these 16 tasks.
    \item \textbf{Compression settings.} We evaluate $p_c \in \{0.25, 0.50, 0.75, 0.90\}$ and set $p_a$ with Eq.~\ref{eq:p_a_deploy}, giving $p_a=\{0.125,0.25,0.125,0.05\}$, respectively.
    \item \textbf{Randomness control.} We use a fixed random seed of 42 for all runs.
    \item \textbf{Inference protocol.} We use KVPress default inference behavior without additional decoding hyperparameter tuning; task-level generation limits follow the benchmark defaults in the KVPress pipeline.
    \item \textbf{Hardware.} Each experiment group can be run on a single NVIDIA H200 GPU.
% [REV-DEL]     \item \textbf{Statistical reporting.} Due to the high computational cost of full-benchmark evaluation, all results are from one full run with fixed seed (no multi-seed repetition and no significance testing).
    \item \rev{\textbf{Statistical reporting.} Greedy decoding makes each run deterministic, so we report one run per setting and quantify uncertainty with a paired bootstrap over test samples (Appendix~\ref{app:significance}).}
    \item \rev{\textbf{Attention implementation.} All reported numbers use PyTorch SDPA attention. VECTOR is also compatible with FlashAttention (Appendix~\ref{app:systems}).}
    \item \rev{\textbf{Code.} Our code is available at \url{https://github.com/yuplin2333/vector-kv-cache-compression}. It includes the VECTOR implementation, the scripts to calibrate $W_\mathrm{OLS}$, and the scripts for all experiments. Our own code is released under the MIT license. Modified files from KVPress and D2O keep their Apache-2.0 license, and modified files from RocketKV keep the NVIDIA License, which permits non-commercial research use only.}
    \item \rev{\textbf{Artifacts and licenses.} We use KVPress (Apache-2.0), D2O (Apache-2.0), RocketKV (NVIDIA License), LongBench (MIT), RULER (Apache-2.0), SCBench (MIT), C4 (ODC-BY), and WikiText-103 (CC BY-SA). The models are Llama-3.1-8B-Instruct (Llama 3.1 Community License), the Qwen3 models (Apache-2.0), and Gemma-3-4B (Gemma Terms of Use). The released code keeps the original licenses of the third-party files it modifies. We use all artifacts for research purposes, consistent with their intended use.}
\end{itemize}

\subsection{NIAH Setup Details}
\begin{itemize}
    \item \textbf{Model and methods.} NIAH is evaluated on \path{meta-llama/Llama-3.1-8B-Instruct} with KeyDiff, SnapKV, KVzip, PyramidKV, and their corresponding VECTOR-augmented variants.
    \item \textbf{Compression setting.} We report the high-compression setting $p_c=0.90$ with $p_a=0.05$ (Eq.~\ref{eq:p_a_deploy}).
    \item \textbf{Grid protocol.} We evaluate a $10 \times 10$ grid of context lengths $\{3000, 6000, \dots, 30000\}$ and needle depths $\{0, 11, 22, \ldots, 100\}\%$.
    \item \textbf{Haystack construction.} Each grid cell uses 5 independent haystacks (\texttt{n\_haystacks}=5) from \texttt{data/NIAH/PaulGrahamEssays}, with the default KVPress NIAH needle/question template.
    \item \textbf{Metric aggregation.} NIAH scores are ROUGE-L F1, averaged over the 5 haystacks per grid cell.
    \item \textbf{Method-specific query awareness.} We follow each method's original setting: \texttt{query\_aware} is enabled for SnapKV/PyramidKV (and their VECTOR variants) and disabled for KeyDiff/KVzip (and their VECTOR variants).
    \item \textbf{Inference and seed.} We use KVPress default inference behavior with fixed seed 42.
\end{itemize}

\rev{\subsection{Approximation Ratio Sweep}
\label{app:pa_sweep}
For the sensitivity study in Section~\ref{sec:experiments:pa_sensitivity}, we use Llama-3.1-8B-Instruct with KeyDiff+VECTOR and KVzip+VECTOR on HotpotQA and NarrativeQA. We sweep $p_a$ from 0 to its upper limit $\min(p_c, 1-p_c)$ in steps of 0.05, at $p_c \in \{0.50, 0.75, 0.90\}$, and report the mean score over the four method--task combinations. All other settings follow the LongBench setup above. Table~\ref{tab:pa_sweep} lists all values shown in Figure~\ref{fig:pa_sensitivity}.}

\begin{table}[t]
\centering
\small
{\color{\revcolor}
\begin{tabular}{cccc}
\toprule
\multirow{2}{*}{$p_a$} & \multicolumn{3}{c}{Compression ratio $p_c$} \\
\cmidrule(lr){2-4}
 & 0.50 & 0.75 & 0.90 \\
\midrule
0.00 & 43.66 & 42.81 & 37.59 \\
0.05 & 43.64 & 43.05 & \textbf{40.35} \\
0.10 & 44.84 & 43.50 & 38.08 \\
0.15 & 44.63 & 43.85 & -- \\
0.20 & 44.32 & \textbf{44.29} & -- \\
0.25 & 44.92 & 41.59 & -- \\
0.30 & 44.85 & -- & -- \\
0.35 & \textbf{45.40} & -- & -- \\
0.40 & 45.14 & -- & -- \\
0.45 & 44.45 & -- & -- \\
0.50 & 41.73 & -- & -- \\
\bottomrule
\end{tabular}
\caption{Mean LongBench score for each approximation ratio $p_a$, averaged over KeyDiff and KVzip on HotpotQA and NarrativeQA (Llama-3.1-8B). Bold marks the best $p_a$ at each compression ratio, and ``--'' marks infeasible values with $p_a > \min(p_c, 1-p_c)$.}
\label{tab:pa_sweep}
}
\end{table}

\rev{\subsection{Reconstructability Signal Ablation}
\label{app:signal_setup}
For the ablation in Section~\ref{sec:experiments:signal}, we use KeyDiff+VECTOR on Llama-3.1-8B-Instruct with six LongBench QA tasks: NarrativeQA, Qasper, MultiFieldQA-en, HotpotQA, 2WikiMQA, and MuSiQue, at $p_c \in \{0.75, 0.90\}$. Both selection rules use the same candidate pool and the same budget, and differ only in how the $2p_a$ approximated tokens are chosen. VECTOR approximates the candidates with the lowest reconstruction error. The importance-ranking rule approximates the least important candidates and retains the most important ones exactly. The correlation between importance and reconstruction error is measured over all candidate tokens on 10\% of the HotpotQA and Qasper samples at $p_c=0.90$, and is 0.166 and 0.164 on the two tasks.}

\section{Ablation Study: OLS vs.\ Analytical Pseudo-Inverse}
\label{app:ols_vs_mp}

We compare the $K \to V$ predictability of the data-driven OLS estimator against the MP pseudo-inverse. For each model, we report the mean and median $R^2$ aggregated across all layers, along with the fraction of layers where MP achieves $R^2 > 0$. OLS results match those in Table~\ref{tab:method_r2}.

\begin{table*}[t]
\centering
% \resizebox{\linewidth}{!}{
\begin{tabular}{lcccccc}
\toprule
& \multicolumn{2}{c}{OLS $R^2$} 
& \multicolumn{3}{c}{MP $R^2$} \\
\cmidrule(lr){2-3} \cmidrule(lr){4-6}
Model & Mean & Median & Mean & Median & Layers with $R^2 > 0$ \\
\midrule
Llama-3.1-8B   & 0.697 & 0.690 & 0.511  & 0.508    & 100\% \\
Qwen3-14B      & 0.695 & 0.691 & -797.0 & -1.545   & 30\%  \\
Qwen3-0.6B     & 0.939 & 0.936 & -9250738 & -2824  & 0\%   \\
Gemma-3-4B     & 0.890 & 0.897 & -3.170 & -0.767   & 29\%  \\
Qwen3-30B-A3B  & 0.686 & 0.720 & -6987  & -572.5   & 0\%   \\
\bottomrule
\end{tabular}
% }
\caption{OLS vs.\ MP pseudo-inverse: layer-aggregated $R^2$ on the held-out validation set. A negative $R^2$ indicates that the predictor performs worse than predicting the mean.}
\label{tab:ols_vs_mp}
\end{table*}

OLS achieves positive $R^2$ on every layer of every model. MP, by contrast, fails systematically. On four of five models, MP produces negative $R^2$ on the majority of layers. On Qwen3-0.6B and Qwen3-30B-A3B, no layer achieves $R^2 > 0$ under MP. Aligning with the theoretical discussion in Section \ref{sec:method}, these results confirm that MP does not minimize the $V$-prediction error. 
% Its failure is not a numerical edge case. It reflects a fundamental mismatch between the MP objective and the target reconstruction task.

\section{Ablation Study: V-only vs.\ K-only Approximation}
\label{app:vonly_vs_konly}

VECTOR adopts a V-only approximation strategy, preserving exact keys and reconstructing values via OLS. A natural alternative is to flip this design: store values exactly and reconstruct keys instead. We refer to this variant as K-only approximation, which trains a V$\to$K OLS estimator and applies forward RoPE to produce approximate keys.

We compare V-only (KeyDiff+VECTOR) and K-only against the KeyDiff baseline on four LongBench tasks using Llama-3.1-8B-Instruct. Both variants use identical budget, importance scorer, and three-way allocation logic, with $p_a$ set by Eq.~\ref{eq:p_a_deploy} as in the main experiments. The only difference is the approximation direction.

\begin{table*}[t]
\centering
% \resizebox{\linewidth}{!}{
\begin{tabular}{llcccc}
\toprule
Method & Comp. Ratio & Qasper & HotpotQA & 2WikiMQA & Musique \\
\midrule
KeyDiff        & 0.25 & 47.30 & 58.39 & 50.69 & 34.08 \\
KeyDiff+VECTOR & 0.25 & 48.18 & 58.45 & 50.72 & 33.59 \\
KeyDiff+K-only & 0.25 & 47.56 & 58.25 & 49.59 & 34.59 \\
\midrule
KeyDiff        & 0.50 & 45.65 & 56.07 & 46.20 & 28.62 \\
KeyDiff+VECTOR & 0.50 & 47.38 & 58.92 & 51.65 & 35.47 \\
KeyDiff+K-only & 0.50 & 43.70 & 54.31 & 49.18 & 30.55 \\
\midrule
KeyDiff        & 0.75 & 31.75 & 52.83 & 38.74 & 24.69 \\
KeyDiff+VECTOR & 0.75 & 39.20 & 54.83 & 45.85 & 28.83 \\
KeyDiff+K-only & 0.75 & 36.59 & 51.74 & 43.52 & 26.24 \\
\midrule
KeyDiff        & 0.90 & 17.00 & 40.03 & 26.18 & 16.01 \\
KeyDiff+VECTOR & 0.90 & 23.96 & 49.88 & 30.72 & 20.72 \\
KeyDiff+K-only & 0.90 & 21.44 & 38.93 & 26.37 & 17.77 \\
\bottomrule
\end{tabular}
% }
\caption{V-only vs.\ K-only approximation on LongBench (Llama-3.1-8B-Instruct). KeyDiff is the unaugmented baseline. KeyDiff+VECTOR approximates values; KeyDiff+K-only approximates keys under the same memory budget.}
\label{tab:vonly_vs_konly}
\end{table*}

Based on Table \ref{tab:vonly_vs_konly}, at $p_c=0.25$, the two variants perform similarly. As compression increases, their behaviors diverge. \rev{V-only approximation improves over the KeyDiff baseline in 15 of the 16 settings, and in all settings with $p_c \geq 0.50$.} K-only approximation shows inconsistent gains. On HotpotQA, it falls below the baseline at $p_c=0.50$, $0.75$, and $0.90$. At $p_c=0.90$, KeyDiff+VECTOR outperforms KeyDiff+K-only by 2.52 points on Qasper and 10.95 points on HotpotQA.

The performance gap is consistent with the asymmetric role of keys and values in attention. Reconstructed keys directly affect the softmax attention weights. Even small errors in keys are amplified nonlinearly by the softmax. This effect compounds under a higher compression ratio. Values, by contrast, enter only the linear weighted sum after softmax. They are substantially more tolerant to approximation error.

\rev{
\section{Competing Methods: Protocol and Full Results}
\label{app:competing}

\paragraph{Common protocol} All competing methods are evaluated on the 16 LongBench tasks with the same prompt construction, generation settings, and scorer as the main experiments. ChunkKV, ExpectedAttention, AdaKV-SnapKV, and TOVA use their KVPress implementations. Following the original setting of each method, ChunkKV, AdaKV-SnapKV, and TOVA compress the context together with the question (query-aware), and ExpectedAttention is query-agnostic. D2O and RocketKV compress during decoding, with the question already in the prompt.

\paragraph{D2O} We use the official implementation of D2O for Llama-3.1-8B. Since the official code supports only Llama models, we port it to Qwen3 for the other two models. The port reproduces the official Llama-3.1-8B scores within 0.77 points on average. We set the sum of the heavy-hitter and recent ratios of D2O to $1-p_c$. D2O scores tokens with the full attention matrix, whose memory grows quadratically with context length. We therefore truncate its contexts to 16K tokens, which keeps 95.5\% of LongBench samples in full. Following the official implementation, D2O caches keys and values per query head after expanding the grouped KV heads. At the same nominal $p_c$, its cache is therefore $4\times$, $5\times$, and $2\times$ larger than that of GQA-preserving methods on Llama-3.1-8B, Qwen3-14B, and Qwen3-0.6B. We keep this behavior unchanged.

\paragraph{RocketKV} We use the official implementation of RocketKV and port it to Qwen3. RocketKV first evicts tokens permanently (Stage 1) and then applies sparse attention over the kept tokens during decoding (Stage 2). We set the Stage-1 budget to $1-p_c$ of the prompt, so that its permanent KV cache matches the other methods, and keep the Stage-2 settings at their defaults. Stage 2 additionally stores chunk statistics and uses a custom sparse-attention kernel. Without compression, the RocketKV pipeline scores 45.90, 50.05, and 30.82 on the three models, which matches our uncompressed baselines within 0.2 points.

\paragraph{Results} Tables~\ref{tab:competing_qwen14b} and~\ref{tab:competing_qwen06b} report results on Qwen3-14B and Qwen3-0.6B. At $p_c=0.90$, SnapKV+VECTOR achieves the best average on both models.}

\begin{table}[t]
\centering
\small
{\color{\revcolor}
\setlength{\tabcolsep}{2.5pt}
\begin{tabular}{lcccc}
\toprule
\multirow{2}{*}{Method} & \multicolumn{4}{c}{Compression ratio $p_c$} \\
\cmidrule(lr){2-5}
 & 0.25 & 0.50 & 0.75 & 0.90 \\
\midrule
No Comp. & \multicolumn{4}{c}{50.08} \\
\midrule
ChunkKV$^\dagger$ & 49.94 & 49.56 & 48.42 & 43.30 \\
ExpectedAttention & 50.28 & \textbf{50.57} & 48.97 & 43.60 \\
AdaKV-SnapKV$^\dagger$ & 49.97 & 49.97 & \textbf{49.21} & 46.91 \\
TOVA$^\dagger$ & 49.93 & 49.69 & 48.98 & 47.18 \\
D2O & 49.39 & 48.83 & 47.05 & 43.88 \\
RocketKV & 49.89 & 49.74 & 49.09 & 46.85 \\
\midrule
KeyDiff + VECTOR & 49.48 & 47.75 & 41.38 & 32.44 \\
SnapKV + VECTOR$^\dagger$ & 49.79 & 49.01 & 48.63 & \textbf{47.65} \\
KVzip + VECTOR & 49.58 & 48.82 & 49.14 & 47.53 \\
PyramidKV + VECTOR$^\dagger$ & \textbf{50.69} & 49.29 & 48.70 & 45.56 \\
\bottomrule
\end{tabular}
\caption{LongBench average on Qwen3-14B for competing methods and VECTOR-augmented methods. Bold marks the best method at each compression ratio. $^\dagger$ denotes query-aware compression.}
\label{tab:competing_qwen14b}
}
\end{table}

\begin{table}[t]
\centering
\small
{\color{\revcolor}
\setlength{\tabcolsep}{2.5pt}
\begin{tabular}{lcccc}
\toprule
\multirow{2}{*}{Method} & \multicolumn{4}{c}{Compression ratio $p_c$} \\
\cmidrule(lr){2-5}
 & 0.25 & 0.50 & 0.75 & 0.90 \\
\midrule
No Comp. & \multicolumn{4}{c}{30.93} \\
\midrule
ChunkKV$^\dagger$ & 31.17 & \textbf{31.18} & 30.65 & 27.93 \\
ExpectedAttention & 30.20 & 29.91 & 22.59 & 18.03 \\
AdaKV-SnapKV$^\dagger$ & 31.04 & 30.71 & 30.59 & 29.84 \\
TOVA$^\dagger$ & \textbf{31.23} & 31.16 & 30.64 & 29.60 \\
D2O & 30.49 & 30.17 & 29.01 & 26.13 \\
RocketKV & 30.45 & 30.38 & 29.34 & 26.94 \\
\midrule
KeyDiff + VECTOR & 30.25 & 28.91 & 22.88 & 18.94 \\
SnapKV + VECTOR$^\dagger$ & 30.99 & 30.47 & 30.96 & \textbf{31.48} \\
KVzip + VECTOR & 30.95 & 29.96 & \textbf{31.45} & 27.05 \\
PyramidKV + VECTOR$^\dagger$ & 26.31 & 26.27 & 25.29 & 24.30 \\
\bottomrule
\end{tabular}
\caption{LongBench average on Qwen3-0.6B for competing methods and VECTOR-augmented methods. Bold marks the best method at each compression ratio. $^\dagger$ denotes query-aware compression.}
\label{tab:competing_qwen06b}
}
\end{table}

\rev{
\section{Implementation and Efficiency Details}
\label{app:systems}

\paragraph{Prefill and decoding} VECTOR makes all routing decisions once at the end of prefill, as its base eviction methods do. The base scorer ranks the tokens, the reconstruction error is computed for the candidate pool, and each token is assigned to a tier. The reconstructed value $\hat{V}=W_\mathrm{OLS}K$ of an approximated token depends only on its retained key. Our implementation computes it in the same pass and writes it into the cache, so decoding is standard attention over the compressed cache. A kernel that stores only keys for approximated tokens can instead reconstruct their values during decoding, which gives the same values.

\paragraph{Memory accounting} Let $B=1-p_c$, and let $C$ denote the memory of the keys (or values) of all tokens. Exactly retained tokens store a key and a value, and approximated tokens store only a key. VECTOR therefore stores $(B-p_a)\cdot 2C + 2p_a \cdot C = 2BC$, which equals the base eviction method. Our implementation is built on KVPress, which keeps keys and values in paired tensors of the same length, and we do not write a custom kernel. Approximated tokens are therefore stored with their reconstructed values, and the measured cache at $p_c=0.90$ is $1.5\times$ that of the base method (Table~\ref{tab:systems_mem}). A key-only layout for approximated tokens attains $2BC$. The OLS matrices add $n_\text{layers} \times d_{kv}^2$ parameters, which is 67\,MB in bf16 for Llama-3.1-8B, independent of context length and batch size.

\paragraph{Timing protocol} We measure on one NVIDIA H200 GPU with batch size 1 and PyTorch SDPA attention, using warm-up runs and CUDA synchronization. Decoding throughput is measured as the difference between a long and a short generation, which cancels the prefill cost. Each configuration is repeated five times, and Table~\ref{tab:systems_decode} reports the mean and standard deviation. These measurements use a PyTorch implementation without custom kernels.

\paragraph{Prefill cost} Value reconstruction adds a nearly constant one-time cost at prefill. At 32K tokens, it is 0.21\,s for Qwen3-14B, 0.18\,s for Llama-3.1-8B, and 0.16\,s for Qwen3-0.6B. Relative to the prefill time, this is 6\% for Qwen3-14B and 9\% for Llama-3.1-8B, and the share decreases as the context grows.

\paragraph{FlashAttention compatibility} VECTOR only rewrites the KV cache after prefill, so it does not depend on the attention kernel. We verified that VECTOR runs with FlashAttention-3~\citep{shah2024flashattention3}. On HotpotQA samples with SnapKV, KeyDiff, and PyramidKV at $p_c=0.90$, 96\% to 98.5\% of the compressed cache entries are identical to the SDPA run, and the remaining differences come from bf16 rounding at the selection boundary. All reported numbers use PyTorch SDPA.
}

\begin{table}[t]
\centering
\small
{\color{\revcolor}
\setlength{\tabcolsep}{3pt}
\resizebox{\columnwidth}{!}{%
\begin{tabular}{lcccc}
\toprule
Model & Full KV & Base & \makecell{+VECTOR\\(ours, impl.)} & \makecell{+VECTOR\\(key-only)} \\
\midrule
Qwen3-14B & 5.000 & 0.500 & 0.751 & 0.500 \\
Llama-3.1-8B & 4.000 & 0.400 & 0.601 & 0.400 \\
Qwen3-0.6B & 3.500 & 0.350 & 0.526 & 0.350 \\
\bottomrule
\end{tabular}}
\caption{KV cache memory (GB) at a 32K context and $p_c=0.90$, for KeyDiff and SnapKV and their VECTOR variants. The first three columns are measured. The last column is the footprint of a key-only layout for approximated tokens ($2BC$).}
\label{tab:systems_mem}
}
\end{table}

\begin{table}[t]
\centering
\small
{\color{\revcolor}
\setlength{\tabcolsep}{4pt}
\resizebox{\columnwidth}{!}{%
\begin{tabular}{lccc}
\toprule
Method & Qwen3-14B & Llama-3.1-8B & Qwen3-0.6B \\
\midrule
KeyDiff & 56.8$\pm$0.3 & 87.4$\pm$2.2 & 93.9$\pm$5.9 \\
KeyDiff + VECTOR & 56.5$\pm$1.2 & 89.9$\pm$4.6 & 104.7$\pm$24.2 \\
\midrule
SnapKV & 56.8$\pm$0.3 & 89.3$\pm$0.5 & 94.0$\pm$3.4 \\
SnapKV + VECTOR & 55.8$\pm$0.7 & 88.3$\pm$2.5 & 91.4$\pm$4.5 \\
\midrule
PyramidKV & 57.0$\pm$0.2 & 89.5$\pm$0.7 & 91.4$\pm$0.7 \\
PyramidKV + VECTOR & 58.2$\pm$2.8 & 88.6$\pm$0.5 & 90.0$\pm$3.8 \\
\bottomrule
\end{tabular}}
\caption{Decoding throughput (tokens/s, mean $\pm$ std over 5 runs) at a 32K context and $p_c=0.90$.}
\label{tab:systems_decode}
}
\end{table}

\rev{
\section{RULER Details}
\label{app:ruler}

We generate RULER data with the official generator, the Llama-3.1-8B-Instruct tokenizer, and seed 42, at 16K, 32K, 64K, and 128K tokens. The RULER score is the mean over 13 subtasks: eight needle-in-a-haystack variants (single, multi-key, multi-value, and multi-query), variable tracking, common and frequent word extraction, and two QA tasks. Each subtask is scored by string match. We use 650 samples at 16K and 32K, 325 at 64K, and 195 at 128K. Only Llama-3.1-8B is evaluated, since it is the only one of our backbones with a context window of at least 128K tokens. SnapKV and PyramidKV and their VECTOR variants are query-aware, as in the main experiments. Table~\ref{tab:ruler_075} reports results at $p_c=0.75$. At 16K, VECTOR improves all four base methods.
}

\begin{table}[t]
\centering
\small
{\color{\revcolor}
\setlength{\tabcolsep}{2.5pt}
\begin{tabular}{lcccc}
\toprule
\multirow{2}{*}{Method} & \multicolumn{4}{c}{Context length} \\
\cmidrule(lr){2-5}
 & 16K & 32K & 64K & 128K \\
\midrule
No Comp. & 94.48 & 88.30 & 85.66 & 80.16 \\
\midrule
KeyDiff & 74.41 & 70.37 & 64.68 & \textbf{61.26} \\
KeyDiff + VECTOR & \textbf{77.38} & \textbf{71.36} & \textbf{66.34} & 59.26 \\
\midrule
SnapKV$^\dagger$ & 82.12 & \textbf{82.80} & \textbf{82.76} & \textbf{78.08} \\
SnapKV + VECTOR$^\dagger$ & \textbf{84.90} & 79.94 & 77.43 & 70.90 \\
\midrule
KVzip & 93.32 & \textbf{88.00} & 83.19 & 72.48 \\
KVzip + VECTOR & \textbf{93.71} & 87.04 & \textbf{84.29} & \textbf{73.91} \\
\midrule
PyramidKV$^\dagger$ & 78.20 & \textbf{78.41} & \textbf{80.65} & \textbf{76.71} \\
PyramidKV + VECTOR$^\dagger$ & \textbf{81.42} & 78.08 & 77.13 & 71.01 \\
\bottomrule
\end{tabular}
\caption{RULER score on Llama-3.1-8B at $p_c=0.75$. Bold marks the better of each base method and its VECTOR variant. $^\dagger$ denotes query-aware compression.}
\label{tab:ruler_075}
}
\end{table}

% [REV] Moved from the main text
\rev{\section{Needle-in-a-Haystack Results}\label{app:niah}}

\begin{figure*}[!t]
    \centering
    \includegraphics[width=0.8\textwidth]{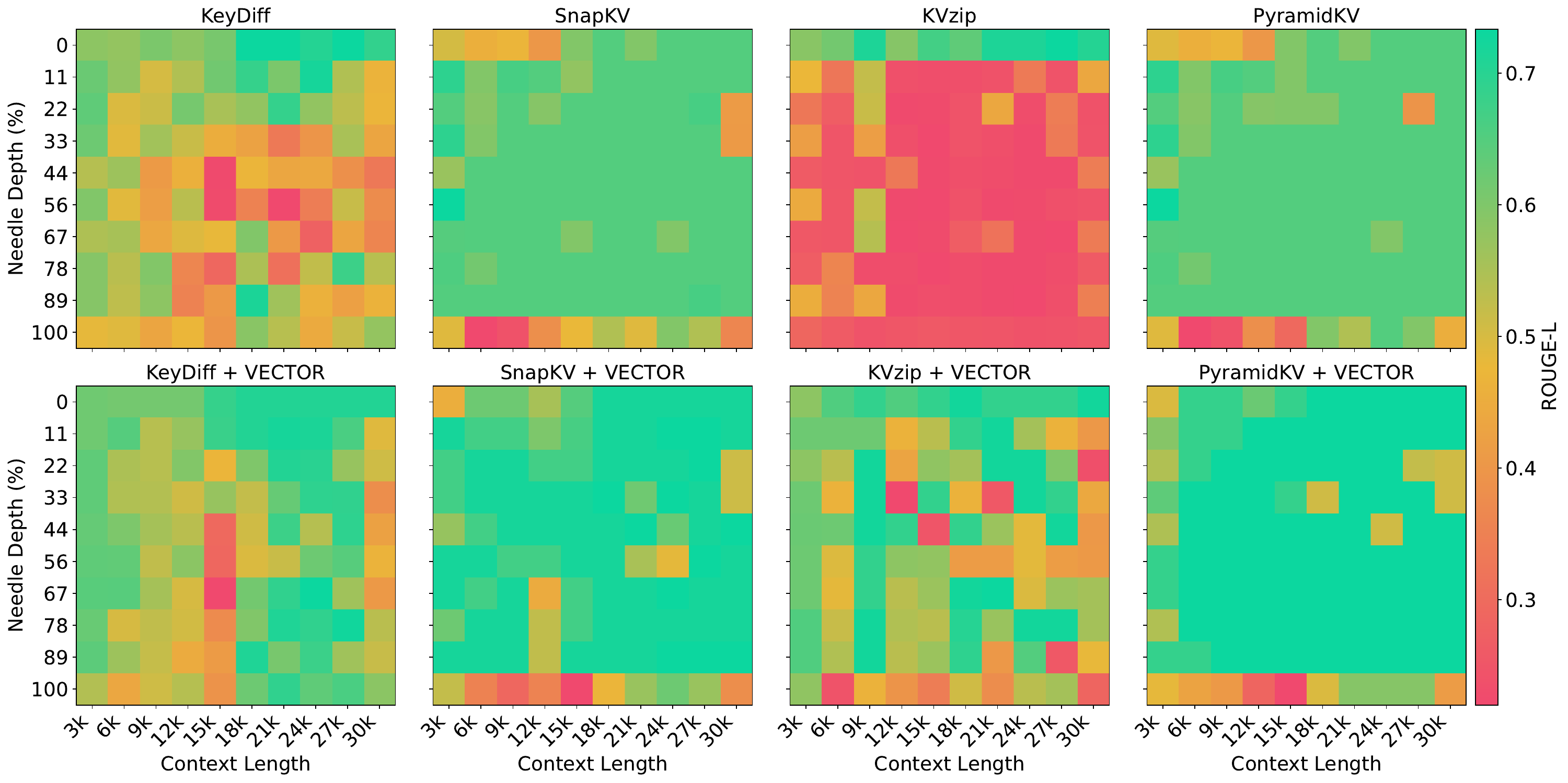}
    \caption{NIAH heatmaps on Llama-3.1-8B at $p_c{=}0.90$.
      \textit{Top row}: KeyDiff, SnapKV, KVzip, and PyramidKV.
      \textit{Bottom row}: corresponding VECTOR-augmented variants.}
    \label{fig:niah}
\end{figure*}

To further assess retrieval robustness under strict memory budgets, we evaluate
NIAH on Llama-3.1-8B across all four baselines and their VECTOR-augmented
variants. We focus on the high-compression regime $p_c{=}0.90$ (with
$p_a{=}0.05$ per Eq.~\eqref{eq:p_a_deploy}), where differences between methods
are most pronounced. We use a $10{\times}10$ grid over context lengths
$\{3\text{k}, 6\text{k}, \ldots, 30\text{k}\}$ and needle depths
$\{0, 11, 22, \ldots, 100\}\%$; each cell is averaged over 5 independent runs
and reported as ROUGE-L F1. Figure~\ref{fig:niah} shows the resulting heatmaps.

Across all four baseline families, VECTOR consistently improves retrieval
quality, though the magnitude of gain varies by baseline. The largest improvement
is observed for KVzip, where the baseline exhibits broad low-score regions and
KVzip+VECTOR substantially recovers performance across both context lengths and
needle depths. For KeyDiff, VECTOR provides clear gains, primarily lifting
low-score cells at harder depth--length combinations. For SnapKV and PyramidKV,
whose baselines are already strong at this compression ratio, improvements are
more moderate but consistently positive\rev{.}

Overall, these results further support the role of reconstructability-aware
approximation under tight memory budgets: rather than merely shifting average
scores upward, VECTOR changes the failure pattern from large contiguous
low-score regions to more localized difficult cells, indicating improved
robustness under challenging retrieval conditions.

\rev{
\section{Multi-Turn Evaluation on SCBench}
\label{app:scbench}

We use SCBench~\citep{li2025scbench} in its shared-context multi-turn setting, with the official prompts in golden-context mode and the official scoring. We use a 128K window with the middle truncation of SCBench. The shared context is compressed once. Every turn is answered from the same compressed cache without recompression, including turns whose questions were not seen during compression. SnapKV and PyramidKV compress the first question together with the context, and KeyDiff compresses the context alone. We pick one task per SCBench category: string retrieval (\texttt{vt}), semantic retrieval (\texttt{choice\_eng}), global information (\texttt{many\_shot}), and multi-tasking (\texttt{summary\_with\_needles}, reported as a summary score and a passkey score). This gives 272 sessions with 1,509 turns. Each task score is the mean over turns, and Avg is the mean of the five columns.

Table~\ref{tab:scbench} reports the results on Llama-3.1-8B. At $p_c=0.90$, VECTOR improves KeyDiff from 52.88 to 57.58 Avg ($+4.69$, 95\% CI $[+2.98, +6.38]$), with the largest gains on string retrieval ($+16.8$) and semantic retrieval ($+11.4$). The gain holds on turns 2 and later, whose questions were never seen at compression time ($+4.52$). Reconstructability is a property of the context and does not depend on the question, so the approximated tokens remain useful for later queries. Compression can raise \texttt{many\_shot} above full KV, since dropping some in-context demonstrations helps this task.
}

\begin{table*}[t]
\centering
\small
{\color{\revcolor}
\resizebox{\textwidth}{!}{%
\begin{tabular}{llccccccc}
\toprule
$p_c$ & Method & vt & choice\_eng & many\_shot & sum. (needles) & passkey (needles) & Avg & Avg (turns $\geq$ 2) \\
\midrule
-- & Full KV & 52.62 & 73.39 & 40.74 & 45.46 & 94.29 & 61.30 & 63.69 \\
\midrule
\multirow{6}{*}{0.75} & SnapKV$^\dagger$ & 51.82 & 69.01 & 42.59 & 43.48 & 93.57 & 60.10 & 61.98 \\
 & SnapKV + VECTOR$^\dagger$ & 47.60 & 69.87 & 42.96 & 43.47 & 90.36 & 58.85 & 60.74 \\
 & KeyDiff & 49.51 & 70.67 & 51.11 & 44.10 & 92.86 & 61.65 & 62.65 \\
 & KeyDiff + VECTOR & 49.87 & 72.06 & 45.18 & 43.19 & 91.79 & 60.42 & 61.02 \\
 & PyramidKV$^\dagger$ & 41.78 & 60.22 & 31.48 & 43.02 & 72.14 & 49.73 & 49.33 \\
 & PyramidKV + VECTOR$^\dagger$ & 40.22 & 60.22 & 33.70 & 41.43 & 72.50 & 49.62 & 49.41 \\
\midrule
\multirow{6}{*}{0.90} & SnapKV$^\dagger$ & 47.02 & 61.99 & 44.81 & 40.23 & 92.86 & 57.38 & 58.23 \\
 & SnapKV + VECTOR$^\dagger$ & 46.80 & 62.40 & 44.07 & 40.78 & 90.36 & 56.88 & 58.17 \\
 & KeyDiff & 27.96 & 51.94 & 54.81 & 41.50 & 88.21 & 52.88 & 53.42 \\
 & KeyDiff + VECTOR & 44.71 & 63.33 & 52.59 & 41.18 & 86.07 & 57.58 & 57.94 \\
 & PyramidKV$^\dagger$ & 37.91 & 56.66 & 33.33 & 39.40 & 68.21 & 47.10 & 45.94 \\
 & PyramidKV + VECTOR$^\dagger$ & 34.44 & 53.14 & 32.59 & 39.22 & 68.21 & 45.52 & 44.37 \\
\bottomrule
\end{tabular}}
\caption{SCBench shared-context multi-turn results on Llama-3.1-8B. Task scores are averaged over all turns. The last column averages only turns 2 and later, whose questions are not seen at compression time. $^\dagger$ denotes query-aware compression.}
\label{tab:scbench}
}
\end{table*}

\rev{
\section{Statistical Significance}
\label{app:significance}

Greedy decoding is deterministic, so runs with different seeds give identical outputs. We therefore quantify uncertainty from the variation over test samples with a paired bootstrap~\citep{koehn2004statistical}. For each comparison, the base method and its VECTOR variant are evaluated on the same 10,000 resamples, and we report the 95\% confidence interval of $\Delta$, the score of the VECTOR variant minus that of the base method. We resample within each task for LongBench and RULER. For SCBench we resample sessions, since all turns in a session share one compressed cache. The LongBench intervals are computed from a rerun of the four Llama-3.1-8B settings that keeps per-sample outputs, and the rerun reproduces Table~\ref{tab:exp_main} within 0.2 points. Tables~\ref{tab:sig_longbench}, \ref{tab:sig_ruler}, and~\ref{tab:sig_scbench} report the results.
}

\begin{table}[t]
\centering
\small
{\color{\revcolor}
\setlength{\tabcolsep}{4pt}
\resizebox{\columnwidth}{!}{%
\begin{tabular}{llccc}
\toprule
$p_c$ & Method & Base & +VECTOR & $\Delta$ [95\% CI] \\
\midrule
0.75 & KeyDiff & 41.68 & 44.79 & $+3.11$ [$+2.38$, $+3.85$] \\
0.75 & SnapKV$^\dagger$ & 45.51 & 45.09 & $-0.43$ [$-1.05$, $+0.19$] \\
0.90 & KeyDiff & 35.24 & 38.94 & $+3.69$ [$+2.93$, $+4.44$] \\
0.90 & SnapKV$^\dagger$ & 43.38 & 45.18 & $+1.80$ [$+1.06$, $+2.54$] \\
\bottomrule
\end{tabular}}
\caption{Paired bootstrap on LongBench (Llama-3.1-8B, 16-task average). $^\dagger$ denotes query-aware compression.}
\label{tab:sig_longbench}
}
\end{table}

\begin{table*}[t]
\centering
\small
{\color{\revcolor}
\resizebox{\textwidth}{!}{%
\begin{tabular}{llcccc}
\toprule
\multirow{2}{*}{$p_c$} & \multirow{2}{*}{Method} & \multicolumn{4}{c}{Context length} \\
\cmidrule(lr){3-6}
 & & 16K & 32K & 64K & 128K \\
\midrule
\multirow{4}{*}{0.75} & KeyDiff & $\mathbf{+2.97}$ [$+1.24$, $+4.76$] & $+0.99$ [$-0.66$, $+2.65$] & $+1.66$ [$-0.18$, $+3.55$] & $-2.00$ [$-5.07$, $+0.98$] \\
 & SnapKV$^\dagger$ & $\mathbf{+2.78}$ [$+1.23$, $+4.30$] & $-2.86$ [$-4.52$, $-1.11$] & $-5.34$ [$-7.45$, $-3.24$] & $-7.19$ [$-10.41$, $-4.11$] \\
 & KVzip & $+0.38$ [$-0.69$, $+1.48$] & $-0.96$ [$-2.29$, $+0.33$] & $+1.10$ [$-1.27$, $+3.51$] & $+1.43$ [$-2.50$, $+5.14$] \\
 & PyramidKV$^\dagger$ & $\mathbf{+3.22}$ [$+1.75$, $+4.63$] & $-0.33$ [$-1.99$, $+1.41$] & $-3.52$ [$-5.99$, $-0.94$] & $-5.70$ [$-9.31$, $-2.24$] \\
\midrule
\multirow{4}{*}{0.90} & KeyDiff & $\mathbf{+1.52}$ [$+0.32$, $+2.81$] & $-0.64$ [$-2.02$, $+0.73$] & $-1.34$ [$-3.33$, $+0.56$] & $+1.52$ [$-1.68$, $+4.77$] \\
 & SnapKV$^\dagger$ & $\mathbf{+12.76}$ [$+10.84$, $+14.61$] & $\mathbf{+5.55}$ [$+3.82$, $+7.29$] & $+2.65$ [$-0.06$, $+5.22$] & $+1.57$ [$-2.66$, $+5.87$] \\
 & KVzip & $\mathbf{+6.22}$ [$+4.43$, $+8.04$] & $\mathbf{+7.64}$ [$+5.89$, $+9.43$] & $\mathbf{+8.31}$ [$+5.24$, $+11.47$] & $\mathbf{+10.54}$ [$+6.21$, $+14.94$] \\
 & PyramidKV$^\dagger$ & $\mathbf{+6.92}$ [$+5.16$, $+8.67$] & $\mathbf{+5.04}$ [$+3.31$, $+6.73$] & $-0.05$ [$-3.02$, $+2.80$] & $+1.99$ [$-2.30$, $+6.48$] \\
\bottomrule
\end{tabular}}
\caption{Paired bootstrap on RULER (Llama-3.1-8B). Each cell reports $\Delta$ (VECTOR variant minus base) and its 95\% confidence interval. Bold marks significant gains. $^\dagger$ denotes query-aware compression.}
\label{tab:sig_ruler}
}
\end{table*}

\begin{table}[t]
\centering
\small
{\color{\revcolor}
\setlength{\tabcolsep}{4pt}
\resizebox{\columnwidth}{!}{%
\begin{tabular}{llccc}
\toprule
$p_c$ & Method & Base & +VECTOR & $\Delta$ [95\% CI] \\
\midrule
0.75 & SnapKV$^\dagger$ & 60.10 & 58.85 & $-1.24$ [$-2.61$, $+0.08$] \\
0.75 & KeyDiff & 61.65 & 60.42 & $-1.23$ [$-2.85$, $+0.40$] \\
0.75 & PyramidKV$^\dagger$ & 49.73 & 49.62 & $-0.11$ [$-1.59$, $+1.35$] \\
0.90 & SnapKV$^\dagger$ & 57.38 & 56.88 & $-0.50$ [$-2.02$, $+0.99$] \\
0.90 & KeyDiff & 52.88 & 57.58 & $+4.69$ [$+2.98$, $+6.38$] \\
0.90 & PyramidKV$^\dagger$ & 47.10 & 45.52 & $-1.58$ [$-3.18$, $+0.00$] \\
\bottomrule
\end{tabular}}
\caption{Paired bootstrap on SCBench Avg (Llama-3.1-8B), resampling sessions. $^\dagger$ denotes query-aware compression.}
\label{tab:sig_scbench}
}
\end{table}

\rev{
\section{Robustness of OLS Calibration}
\label{app:calib}

We vary only the calibration data of $W_\mathrm{OLS}$ for Llama-3.1-8B, and keep all other settings unchanged. We replace C4 with WikiText-103~\citep{merity2017pointer} using the same number of sequences (5,000), and we reduce the C4 calibration set by $10\times$ to 500 sequences. Table~\ref{tab:calib} reports LongBench averages for KeyDiff+VECTOR and SnapKV+VECTOR. None of the eight differences from the default calibration is significant under the paired bootstrap. The largest change is 0.32 points, which is an order of magnitude below the gain of VECTOR over its base method. Calibration on 500 sequences takes about five minutes on one GPU.
}

\begin{table*}[t]
\centering
\small
{\color{\revcolor}
\resizebox{\textwidth}{!}{%
\begin{tabular}{lcccc}
\toprule
Calibration data & KeyDiff+VECTOR, $p_c{=}0.75$ & KeyDiff+VECTOR, $p_c{=}0.90$ & SnapKV+VECTOR$^\dagger$, $p_c{=}0.75$ & SnapKV+VECTOR$^\dagger$, $p_c{=}0.90$ \\
\midrule
C4, 5,000 seq.\ (default) & 44.79 & 38.94 & 45.09 & 45.18 \\
WikiText-103, 5,000 seq. & 44.47 ($-0.32$) & 38.86 ($-0.07$) & 44.98 ($-0.11$) & 45.38 ($+0.20$) \\
C4, 500 seq. & 44.51 ($-0.28$) & 38.96 ($+0.02$) & 45.17 ($+0.09$) & 45.09 ($-0.09$) \\
\midrule
Base method (no VECTOR) & 41.68 & 35.24 & 45.51 & 43.38 \\
\bottomrule
\end{tabular}}
\caption{LongBench average on Llama-3.1-8B with different calibration data for $W_\mathrm{OLS}$. Values in parentheses are differences from the default calibration. All runs use the rerun setting of Appendix~\ref{app:significance}. $^\dagger$ denotes query-aware compression.}
\label{tab:calib}
}
\end{table*}

\end{document}